\documentclass{article}

\usepackage{amsmath,amsfonts,bm}

\def\eqref#1{equation~\ref{#1}}

\def\1{\bm{1}}

\DeclareMathAlphabet{\mathsfit}{\encodingdefault}{\sfdefault}{m}{sl}
\SetMathAlphabet{\mathsfit}{bold}{\encodingdefault}{\sfdefault}{bx}{n}

\usepackage{microtype}
\usepackage{graphicx}
\usepackage{subcaption}
\usepackage{booktabs} 

\usepackage{hyperref}

\usepackage[accepted]{icml2024}

\usepackage{amsmath}
\usepackage{amssymb}
\usepackage{mathtools}
\usepackage{amsthm}

\usepackage{enumitem}

\usepackage[capitalize,noabbrev]{cleveref}

\theoremstyle{plain}
\newtheorem{theorem}{Theorem}[section]
\newtheorem{proposition}[theorem]{Proposition}
\newtheorem{lemma}[theorem]{Lemma}
\newtheorem{corollary}[theorem]{Corollary}
\theoremstyle{definition}
\newtheorem{notation}{Notation}[section]
\newtheorem{definition}[theorem]{Definition}
\newtheorem{assumption}[theorem]{Assumption}
\theoremstyle{remark}
\newtheorem{remark}[theorem]{Remark}

\usepackage[textsize=tiny]{todonotes}

\newcommand{\cO}{\mathcal O}
\newcommand{\cL}{\mathcal L}
\newcommand{\cU}{\mathcal U}
\newcommand{\bbR}{\mathbb R}

\newcommand{\pp}[2]{\frac{\partial#1}{\partial#2}}

\renewcommand{\vec}[1]{\mathbf{#1}}
\renewcommand{\footnotesize}{\fontsize{8pt}{11pt}\selectfont}

\icmltitlerunning{Efficient multirate gradient descent schemes for data-induced multiscale losses}

\begin{document}
\onecolumn
\icmltitle{Data-induced multiscale losses and efficient multirate gradient descent schemes}



\icmlsetsymbol{equal}{*}

\begin{icmlauthorlist}
\icmlauthor{Juncai He}{kaust}
\icmlauthor{Liangchen Liu}{austin}
\icmlauthor{Richard Tsai}{austin,oden}
\end{icmlauthorlist}

\icmlaffiliation{kaust}{Computer, Electrical and Mathematical Science and Engineering Division, King Abdullah University of Science and Technology, Thuwal, Saudi Arabia}
\icmlaffiliation{austin}{Department of Mathematics, University of Texas at Austin,
Austin, TX, USA}
\icmlaffiliation{oden}{Oden Institute for
Computational Engineering and Sciences, University of Texas at
Austin, Austin, TX, USA.}


\icmlkeywords{Multiscale data, multiscale loss, gradient descent, learning rate schedule}

\vskip 0.3in



\printAffiliationsAndNotice{}  

\begin{abstract}
This paper investigates the impact of multiscale data on machine learning algorithms, particularly in the context of deep learning. A dataset is multiscale if its distribution shows large variations in scale across different directions. This paper reveals multiscale structures in the loss landscape, including its gradients and Hessians inherited from the data.
Correspondingly, it introduces a novel gradient descent approach, drawing inspiration from multiscale algorithms used in scientific computing. 
This approach seeks to transcend empirical learning rate selection, offering a more systematic, data-informed strategy to enhance training efficiency, especially in the later stages.
\end{abstract}

\section{Introduction}

In many supervised learning setups, the input data are commonly embedded in high-dimensional Euclidean spaces for convenience. To clarify, we are referring to the non-label part of the datasets (i.e. the features). A natural question arises concerning the distributions of the training data in the embedding space and their potential consequences on the efficacy of learning. 
A substantial body of literature~\cite{niyogi2008finding, peyre2009manifold, carlsson2009topology} has delved into the ``manifold hypothesis", which suggests the data distributions behind the complex representations might concentrate near a simpler, lower-dimensional manifold. Empirical validations~\cite{scholkopf1998nonlinear, brand2002charting, saul2003think} and the development of the theoretical testing framework~\cite{narayanan2010sample,fefferman2016testing} have lent support to such hypothesis, and recent research~\cite{medina2019heuristic,brown2022verifying} has explored its variants.

The manifold hypothesis has sparked exciting works and numerous insights discovered by the manifold learning community~\cite{roweis2000nonlinear, tenenbaum2000global, belkin2003laplacian, donoho2003hessian, weinberger2006unsupervised}. More recently, it has found applications in the realm of deep learning in various areas, including approximation theory~\cite{chen2019efficient,cloninger2021deep,schonsheck2022semi}, intrinsic dimensions~\cite{brahma2015deep,ansuini2019intrinsic, pope2021intrinsic}, initialization~\cite{tiwari2022effects}, and the effects of training ~\cite{he2023side}.

  



Nevertheless, the predominant techniques and analysis in deep learning implicitly assume the distribution of features is full-dimensional in the embedding space and that it has the same scale across different directions. This assumption is evident in the prevalent initialization methods~\cite{glorot2010understanding,he2015delving} for parameters, including those acting on the input data.

However, even under the simplistic model of principal component analysis (PCA)~\cite{pearson1901liii,hotelling1933analysis}, many real-world datasets exhibit a fast-decay spectrum in their principal directions, suggesting data distributions across different directions can manifest drastically different scales. We refer to datasets with variations in the orders of magnitude of scales across different principal directions as \emph{multiscale datasets}, prompting a pertinent question: How do multiscale datasets impact algorithms, particularly those heavily reliant on data, such as gradient descent (GD) in deep learning?
This article aims to unveil the potential impacts of multiscale data on machine learning algorithms, with a specific focus on deep learning problems. Notably, the multiscale nature of data passes down to the learning problem, giving rise to a multiscale loss landscape from various perspectives. 

Stochastic gradient descent (SGD) algorithms with varying learning rates \cite{robbins1951stochastic,you2019does,smith2019super,liu2020variance} have been instrumental in the success of deep learning tasks. While it is widely believed that adopting specific learning rate schedules can significantly enhance the generalization ability of deep neural networks (DNNs), such learning rate schemes have primarily relied on empirical evidence and arguments. This suggests that there is still room for further understanding and improvement of SGD-based algorithms when applied to real-world multiscale datasets. 

In the scientific computing community, 
multiscale algorithms such as \cite{engquist2005heterogeneous, tao2010nonintrusive} can efficiently compute dynamical systems where they leverage the fast averaging and near independence in components due to the gap in two widely separated scales in the systems. Drawing inspiration from these concepts, we propose a novel yet simple explicit gradient descent scheme that adopts learning rates based on multiscale information from the data. The objective is to bridge the gap between the empirical nature of learning rate choices and a more systematic, multiscale approach, ultimately aiming for the efficiency and adaptability of training algorithms for models defined on real-world multiscale datasets.
While our focus doesn't extend to addressing SGD's selection of minima, we argue that the proposed algorithm is suitable for accelerating the later/final stages of gradient-based training of DNNs.

Our main contributions include:
\begin{enumerate}
    \item We prove that multiscale data and $\ell^2$ loss lead to multiscale landscapes. Specifically, we derive a multiscale expansion of the loss gradients and reveal component-wise multiscale characteristics within gradients and Hessians. We verify these theoretical findings by numerically investigating DNN models on the CIFAR dataset.
    \item We propose an explicit multirate gradient descent (MrGD) scheme that leverages multiscale information to construct an appropriate learning rate schedule for much-enhanced convergence speed.
    \item We establish a comprehensive and rigorous theory demonstrating that the MrGD scheme achieves a quasi-optimal convergence rate for linear problems and can be extended to convex functions.
\end{enumerate}


\subsection{Related Work}
\paragraph{Multiscale landscape of loss in learning.}
There has been considerable discussion about the origins of the multiscale expansion form in the loss gradient. \cite{mei2018landscape} suggests a possible origin of the loss with a multiscale structure arising from the effect of noise in various statistical models. \cite{kong2020stochasticity} demonstrates that a simple 2-layer neural network (NN) using a periodic activation function trained with multiscale data leads to a loss exhibiting a multiscale expansion. They also show that deterministic gradient descent can become stochastic with a sufficiently large learning rate. \cite{ma2022beyond} numerically observe a multiscale behavior in the loss landscape of neural network loss functions, which is manifested in two ways: (1) in the vicinity of minima, the loss combines a continuum of scales and grows sub-quadratically, and (2) over a larger region, the loss displays several distinct scales. Recently, \cite{he2023side} analyzed linear regression on two-scale data sampled from a distribution that concentrates around a lower dimensional linear subspace. The regression solutions reflect scales corresponding to the underlying data. Their results also reveal a non-trivial relationship between the number of data points and the magnitude of the smaller scale, which is crucial for the stability of the linear model, a concept that extends to deep neural networks. \cite{liu2023linear} extended the data distributions to embedded curved manifolds, providing closed-form solutions for local linear regression on hypersurfaces and curves, indicating a potential influence of the multiscale geometric properties of the underlying data manifold on the regression solutions.

\paragraph{Larger learning rates in GD and SGD}
Several studies, both numerical~\cite{loshchilov2017sgdr,you2019does,smith2019super,liu2020variance} and theoretical~\cite{oymak2021provable,wang2023convergence,das2023branch,grimmer2023provably}, have shown that an appropriately larger learning rate can benefit the training of GD and SGD in general convex optimization and deep learning. However, most of these studies focus on improvements for better generalization, such as Warm Restart~\cite{loshchilov2017sgdr} for training general NNs towards flat local minima~\cite{he2019asymmetric}, or for specific problems like matrix factorization~\cite{wang2022large} and deterministic and stochastic Hamiltonian dynamical systems~\cite{li2023nysalt}. In this work, we propose a multiscale landscape perspective of the empirical loss and demonstrate that appropriately larger learning rates may also contribute to improved convergence during training.

\section{The loss landscape}
In this section, we discuss the empirical loss landscape arising from multiscale data.  We first make a key assumption on the data in the following:
\begin{assumption}\label{assumption:1} 
The dataset,  $\left\{({\vec x_i}, g(\vec x_i))\right\}_{i=1}^N \subset \mathbb R^{d+1}$, comprises $N$ $i.i.d.$ samples $\{\vec x_i\}_{i=1}^N\subset \bbR^d$ drawn from distributions with a \emph{component-wise multiscale structure}. This structure signifies the distribution scales along different directions can be organized into $(m+1)$ groups based on their magnitude, where $\sum_{i=0}^m d_i = d$. More specifically, there exists a unitary matrix $UU^T=I_{d\times d}$ $s.t.$
    \begin{equation}
    \widetilde{\vec x}_i = U^T({\vec x_i}-\bar{\vec x}), 
\end{equation}
where $ \widetilde{\vec x}_i$ satisfies:
\begin{equation}\label{eq:multi-data}
    \widetilde{\vec x}_i = \left(\widetilde{\vec x}^0_i, \,\varepsilon_1 \widetilde{\vec x}^1_i, \, \dots,\, \varepsilon_m \widetilde{\vec x}^m_i \right)^T \in \mathbb R^{d},
\end{equation}
with $1 \gg \varepsilon_1\gg\varepsilon_2\gg\dots\gg\varepsilon_m>0$. Here, $\vec x^k_i \sim \mathcal U_k$ on $\mathbb R^{d_k}$ for $k=0:m$, each representing a group of directions of the $k^{th}$-scale, where $\cU_k$ is the corresponding uniform distribution with a scale of $\cO(1)$ in all directions. 
\end{assumption}

In practice, PCA can be used to transform data into the above form. Thus, for the convenience of analysis, we shall assume the presence of such preprocessing and identify $\widetilde{\vec x}_i$ with $\vec x_i$, i.e. $U=I$ and $\bar{\vec x}=\bar 0$.

\subsection{Logistic regression with multiscale data}\label{sec:log_regress_multiscale}

We consider logistic regression for classifying the input vector $\vec x$ into $k$-classes, with a learning function $f(\vec x; W):\mathbb{R}^d\mapsto \mathbb{R}^k$, 
    \begin{equation}\label{eq:logit}
        \left[f({\vec  x; \vec w, b})\right]_j  = \frac{e^{\vec w_j \cdot \vec  x + b_j}}{\sum_{i=1}^k e^{\vec w_i \cdot \vec  x + b_i}}, \quad j = 1:k,
    \end{equation}
and 
the cross-entropy loss
    \begin{equation}\label{eqn:cross_entropy}
    \mathcal L_{c}(W) =  -\frac{1}{N} \sum_{i=1}^N  g(\vec x_i)\cdot\log f\left( \vec x_i;W \right).
\end{equation}
The subsequent proposition illustrates the emergence of multiscale gradient components due to multiscale data:
\begin{proposition}\label{thm:grad_logit}
When the data $\{\vec x_i\}_{i=1}^N$ satisfies~\Cref{assumption:1}, the loss in ~(\ref{eqn:cross_entropy}) has a multiscale gradient component: 
    \begin{equation}\label{eqn:multiscale_grad}
    \frac{\partial \mathcal L_c}{\partial \vec w} \sim \big.\left(\cO(1), \cO(\varepsilon_1), \cO(\varepsilon_2)\dots \big.\right),
\end{equation}
where $\cO(\varepsilon_j)$ is a group of vectors in $\mathbb{R}^k$ of magnitudes bounded above by  a constant multiple of $\varepsilon_j$.
\begin{proof}
    See~\Cref{proof:multiscale_grad_logit}.
\end{proof}
\end{proposition}


We remark that an identical form of the gradient also arises in linear regression problem with the least square loss.

\subsection{Deep learning with multiscale data}\label{sec:NN_multiscale}


We consider neural network function classes defined by
    \begin{equation}\label{eq:defdnn}
        \begin{cases}
            f^0(\vec  x) &= \vec  x \\
            f^\ell(\vec  x; W) &= \sigma\left(W^\ell f^{\ell-1}(\vec  x) + b^\ell \right), \;\,\ell=1:L \\
            f({\vec  x; W}) &= W^{L+1}f^L(\vec x)
        \end{cases},
    \end{equation}
    where $\sigma$ is the activation function. 
The entries in the matrix $W^\ell$ will be denoted by $W^\ell_{i,j}$.
We will use the simple least square loss
\begin{equation}\label{eqn:least_sqaure}
    \mathcal L(W) = \frac{1}{2N} \sum_{i=1}^N \left.\big(f({\vec x_i;
W}) - g_i\right.\big)^2.
\end{equation}

\paragraph{Multiscale gradient components}
Due to the multiplicative structure in the network's first layer,
the neural networks defined in~(\ref{eq:defdnn}) can be represented as $f(x,W) = \tilde{f}(W^1x; W^2)$. 
This form is reminiscent of those from the regression models~(\ref{eq:logit}), indicating a comparable multiscale effect on $W_1$, similar to~\Cref{thm:grad_logit}.

\begin{figure}[ht]
\begin{center}
\hspace*{3em}
\begin{subfigure}[b]{0.36\linewidth}
            \centering
            \includegraphics[width=\textwidth]{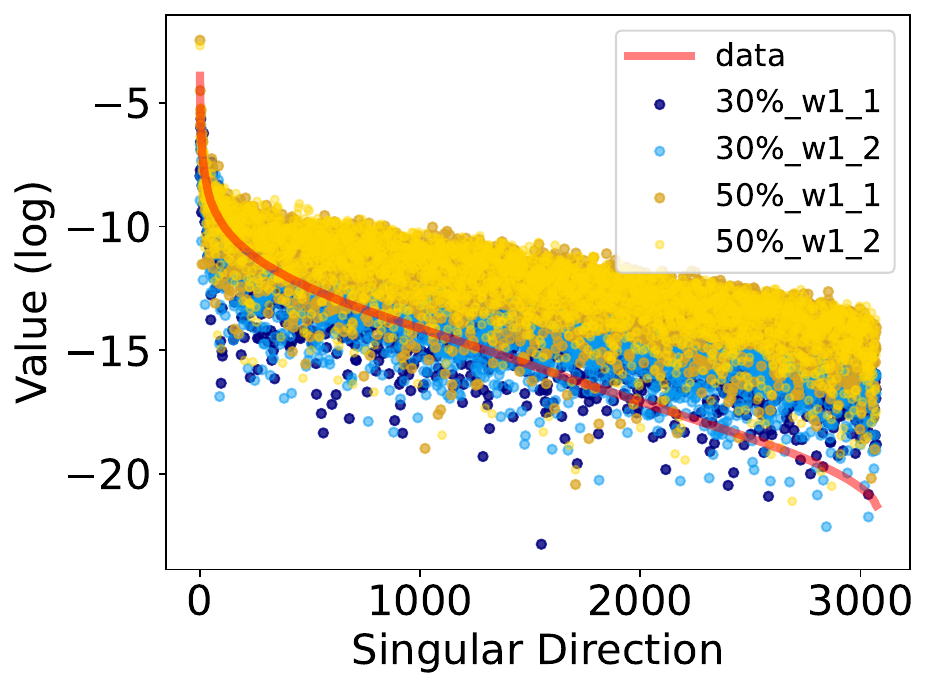}
            \caption[]%
            {First layer; original order}
            \label{fig:cifar_grad_w1_unorder}
        \end{subfigure}\hspace*{\fill}
        \begin{subfigure}[b]{0.36\linewidth}  
            \centering 
            \includegraphics[width=\textwidth]{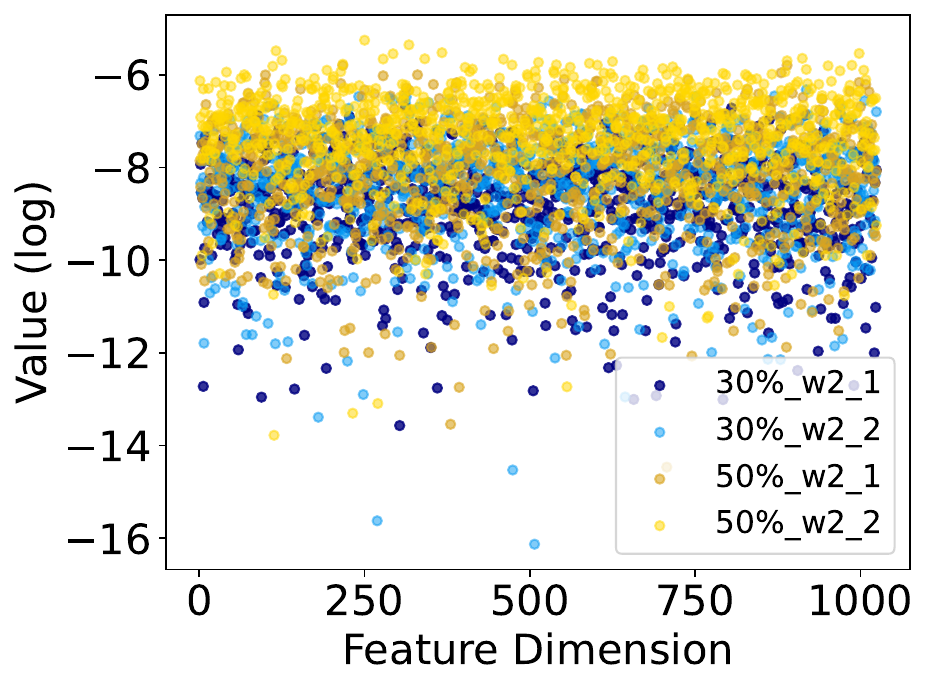}
            \caption[]%
            {Second layer; original order}
            \label{fig:cifar_grad_w2_unorder}
        \end{subfigure}\hspace*{\fill}
        \\
        \hspace*{3em}
        \begin{subfigure}[b]{0.36\linewidth}
            \centering
            \includegraphics[width=\textwidth]{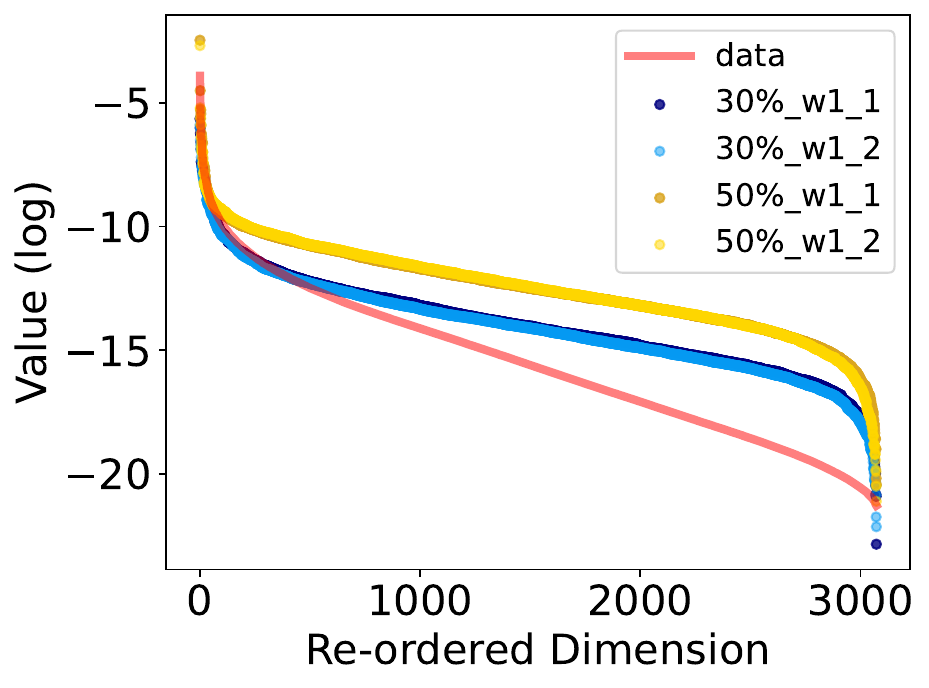}
            \caption[]%
            {First layer; re-ordered}    
            \label{fig:cifar_grad_w1_order}
        \end{subfigure}\hspace*{\fill}
        \begin{subfigure}[b]{0.36\linewidth}  
            \centering 
            \includegraphics[width=\textwidth]{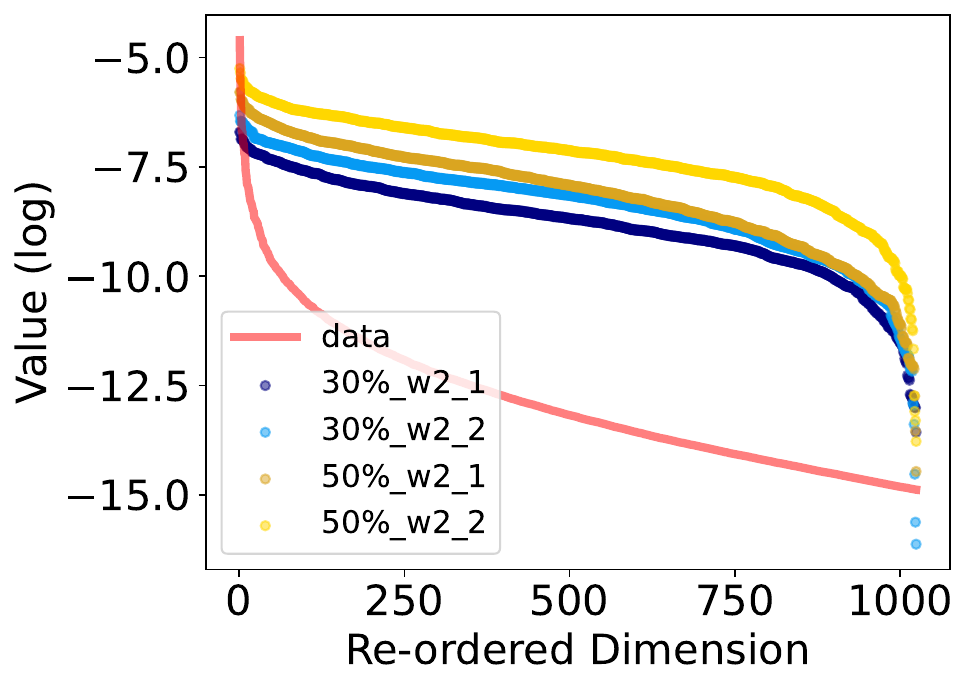}
            \caption[]%
            {Second layer; re-ordered }    
            \label{fig:cifar_grad_w2_order}
        \end{subfigure}\hspace*{\fill}
\end{center}
\caption{The magnitudes of the loss gradient with respect to $W^1_{1,:}$ and $W^1_{i,:}$ for a randomly chosen $i>0$ on the natural log scale; the quantities are labelled \texttt{w1\_1} and \texttt{w1\_2} respectively. 
Two models with test accuracy $30\%$ (blue) and $50\%$ (yellow) are investigated. The bottom row has gradient values sorted by their magnitudes. The red curve represents the principal values of the training data, scaled to align with the gradient magnitude for better comparison. The red curve is truncated in the bottom-right.}
\label{fig:cifar_grad}
\end{figure}

\begin{proposition}\label{thm:grad_NN}
    Suppose $f^1\in \mathbb{R}^{n_1}$.
    When the data $\{\vec x_i\}_{i=1}^N$ satisfies~\Cref{assumption:1}, the loss~(\ref{eqn:least_sqaure}) has:
    \begin{equation}
    \frac{\partial \mathcal L}{\partial W^1_{i,:}} \sim \big.\left(\cO(1), \cO(\varepsilon_1), \cO(\varepsilon_2)\dots \big.\right),~~~1\le i\le n_1, 
\end{equation}
where $\cO(\varepsilon_i)$ are components in the gradient of order $\varepsilon_i$.
\begin{proof}
    See~\Cref{proof:multiscale_grad_NN}.
\end{proof}
\end{proposition}
We conduct the following experiments to verify the proposition empirically. We train a simple $3$-layer multi-layer perceptron (MLP) of sizes $3072$-$1024$-$128$-$10$ using full gradient descent with the cross-entropy loss on the CIFAR10 dataset~\cite{krizhevsky2009learning}. 
The dataset is pre-aligned using PCA, as described in~\Cref{assumption:1}. 
We examine the magnitude of 
$\partial\mathcal{L}/\partial{W^1_{i,:}}$
for $i=1$ and another randomly chosen index. \Cref{fig:cifar_grad} shows these quantities 
at different stages of training, corresponding to test accuracies around $30\%$ and $50\%$. (Trained $3$-layer MLPs typically reach a test accuracy cap of around $60\%$ for CIFAR.) We then compare the gradients across the first and second hidden layers.

From~\Cref{fig:cifar_grad}, it is evident that the raw gradient magnitudes of the first hidden layer closely track the decay of the scale of the data distribution. In contrast, the gradient in the second hidden layer exhibits a more uniformly scattered pattern. This reconfirms
the fact that the gradient in the first layer is primarily characterized by the data.


If the data exhibits a power cascade of scales, one can further derive a multiscale expansion of the loss gradient with respect to \emph{all} weights in the network.

\paragraph{Multiscale expansion of the loss gradient}
We consdier a special case of~(\ref{eq:multi-data}), in which
\begin{equation}\label{eqn:multi-data_distinct}
    \vec x_i = \left(\vec x^0_i, \,\varepsilon \vec x^1_i, \varepsilon^2 \vec x^2_i\, \dots,\, \varepsilon^m \vec x^m_i \right)^T,
\end{equation}
for some $\varepsilon\ll1$.  We establish in the following Theorem: 
\begin{theorem}\label{thm:NN_loss_expand}
For any functions defined in~(\ref{eq:defdnn}), the loss function~(\ref{eqn:least_sqaure}) with a dataset under Assumption~\ref{assumption:1} along with~(\ref{eqn:multi-data_distinct}):
\begin{equation}\label{eq:multiscale-expansion}
    \frac{\partial \mathcal L}{\partial W^\ell} = \frac{1}{N}\sum_{k=0}^m \varepsilon^k A^\ell_k\left(\vec x^0,\cdots, \vec x^k \right),
\end{equation}
\begin{equation}
    A^{\ell}_0(\vec x^0) = \left(f(\vec x^0)-g(\vec x^0)\right)\widetilde \lambda^\ell(\vec x^0) \left(f^{\ell-1}(\vec x^0)\right)^T,
\end{equation}
where $\widetilde  \lambda^\ell(\vec x^0)$ is a tensor product related to $\partial f/\partial f^l$, given in~\Cref{eqn:lambda_tilde}, and $A^\ell_k\left(\vec x^0,\cdots, \vec x^k \right)$ is of $\cO(1)$ defined in~\Cref{eq:Akxk}.

\begin{proof}
    Apply induction on each component of the loss gradient. For details, see~\Cref{sec:proof_dnn_expansion}.
\end{proof}
\end{theorem}

Notice that in the power series, $A^\ell_k$ depends \emph{only} on the first $k+1$ coordinates.
Therefore, the empirical loss is relatively insensitive to the smaller scale coordinates, $\vec x^\nu, \nu=2,3,\cdots.$ Consequently, relying solely on monitoring the training loss may not provide a sufficiently accurate indication of the inference quality, especially for inputs close to the tail of the multiscale data distribution.



With such an additive multiscale structure of the loss gradient, \cite{kong2020stochasticity} show that solutions of the deterministic GD may exhibit stochastic behavior. However, this observation is contingent upon an assumption that the small scales are periodic in a weak sense. Analyzing the loss gradient with all scales mixed up is, in general, challenging and beyond the scope of this work.

\paragraph{Multiscale Hessian}
A multitude of studies have utilized local quadratic approximations and second-order information of the loss to scrutinize the training dynamics of deep neural networks~\cite{wu2018sgd, du2018gradient, jacot2018neural}. In these investigation, the convergence behavior of the neural network and the selection of minima are typically influenced by the eigenvalues of the Hessian. On the other hand, of particular concern to us is the learning rate schedule for the gradient-based training of neural networks. Inevitably, the learning rate is limited by the stability of the SGD, a constraint determined by the eigenvalues of the loss's Hessian. Thus, the characteristics of the Hessian play a crucial role in unraveling many underlying issues.

Following the previous derivation, we present a result demonstrating the connections between the data's characteristics and the Hessian of the weights related to the first layer, denoted by $\nabla^2_{W^1_{i,:}}$:
\begin{proposition}\label{thm:Hess}
    Let $f^1\in\mathbb{R}^{n_1}.$ Then for any $1\le i\le n_1$
    \[ \nabla^2_{W^1_{i,:}} \cL  = \frac{1}{2N}\sum_{j=1}^{N} s_i(\vec x_j) \vec x_j\vec x_j^T \approx \left(\sum_{j=1}^N s_i(\vec x_j)\right) X^TX ,\]
    where, denoting the first argument of $\tilde f$ as $\vec z = (z_1, z_2, \dots, z_{n_1})$,
\[s_i(\vec x_j) := c_i(\vec x_j)^2+ \left( f(\vec x_j, W) -g_j \right)    
    \pp{^2 \tilde{f}(\vec z, W^2)}{z_i^2}\in\bbR,
{(W^1_{i,:}\vec x_j)^2}\in\bbR. \qquad c_{i}(\vec x_j):=  \left.\pp{\tilde f(\vec z, W^2)}{ z_i}\right. \in \bbR\]
    
    for $\vec z=W^1\vec x_j$.
    \begin{proof}
    See~\Cref{proof:multiscale_Hess_NN}.
    \end{proof}
    
\end{proposition}

Note $X^TX$ is the (sample) covariance matrix. Therefore, the spectral distribution of the Hessian $\nabla^2_{W^1_{i,:}} \cL$ at any stage of training should also resemble that of the data, up to a constant multiplier. The same analysis applies to $\nabla_{W^1_{s,:}}\nabla_{W^1_{t,:}} \cL$ for any $1\le s,t \le n_1$.

\begin{figure}[ht]
\vskip 0.2in
\begin{center}
\hspace*{3em}
\begin{subfigure}[b]{0.36\linewidth}
            \centering
            \includegraphics[width=\textwidth]{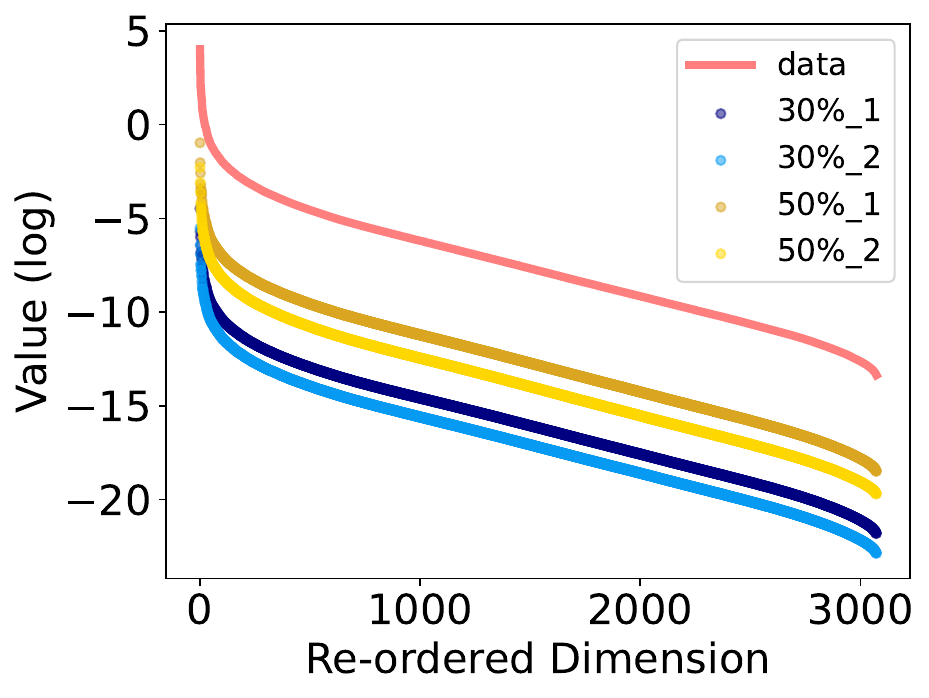}
            \caption[]%
            {First layer}    
        \end{subfigure}
        \hspace*{\fill}
        \begin{subfigure}[b]{0.36\linewidth}  
            \centering 
            \includegraphics[width=\textwidth]{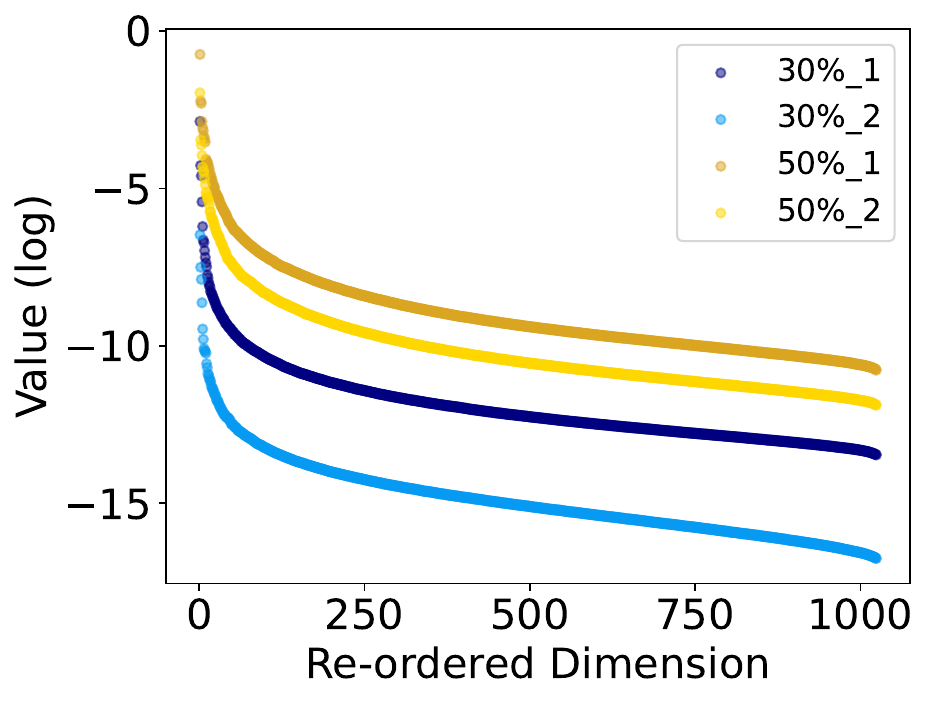}
            \caption[]%
            {Second layer}    
        \end{subfigure}
        \hspace*{\fill}
\caption{Eigenvalues of the Hessians, $\nabla^2_{W^1_{i,:}}\mathcal{L}$ and $\nabla^2_{W^2_{i,:}}\mathcal{L}$, on the natural log scale, under the same setup as in \Cref{fig:cifar_grad}. The eigenvalues are sorted by magnitude. The original eigenvalues (not shown) only exhibit slight jittering toward the ending dimensins, with overall trends closely mirroring the ordered ones.}
\label{figs:cifar_hess}
\end{center}
\vskip -0.2in
\end{figure}

We also confirm this observation through the same numerical experiments setups as before, and present the results in \Cref{figs:cifar_hess}. We notice that, at any stage of training, the spectrum of the Hessian for the first layer displays a decay trend identical to the spectrum of the data. Notably, the spectrum of the Hessian for the second layer also demonstrates a rapid decay in its leading components. We conjecture this phenomenon is related to~\Cref{thm:NN_loss_expand}. This phenomenon is also evident in the training of MNIST (refer to~\Cref{figs:cifar_mnist_w2}).

Extending a parallel derivation to previous regression models also leads to a Hessian characterized by the data covariance matrix, due to the structural resemblance between these models. 
The above observations bring forth the question:
how can one leverage such information to devise gradient descent algorithms that effectively exploit the inherent multiscale characteristics identified in the data?
\section{Multirate gradient descent}

In this section, we introduce and analyze a novel gradient descent algorithm, tailored for the cases discussed in the previous section. It will become evident that by a blend of large and small learning rates, the algorithm  capitalizes on the multiscale characteristics, resulting in accelerated convergence.

\subsection{Multirate gradient descent for quadratic problems}\label{sec:mr_linear}
Consider the following minimization problem:
\begin{equation}\label{eq:A}
    \min_{\theta \in \mathbb R^d} \mathcal L(\theta) :=  \frac{1}{2} \theta^T A \theta - g^T\theta,
\end{equation}
where $A$ is a $d\times d$ symmetric matrix with strictly positive eigenvalues, and the eigenvalues can be divided into $m$ groups, each comprising $d_i$ members such that $d=\sum_i^m d_i$, following~\Cref{assumption:1} for data. 
These eigenvalues are arranged in non-increasing order as:
\begin{equation}\label{eqn:spectrum}
    \sigma_{1,1} \ge \sigma_{1,2} \ge \cdots \ge \sigma_{1,d_1} > \sigma_{2,1} \ge \cdots \ge \sigma_{2,d_2} > \cdots >  \sigma_{m,1} \ge \cdots \ge \sigma_{m,d_m}
\end{equation}

For $i=1:m$, denote $\sigma_i := \sigma_{i,1}$ the principal eigenvalue and $\kappa_i := \displaystyle\frac{\sigma_{i,1}}{\sigma_{i,d_i}}$ the local condition number for group $i$.
Additionally, we introduce the definition of decay rate:
\begin{definition}
    The \emph{decay rate} between consecutive groups of eigenvalues is defined by:
    \[ r_i := \frac{\sigma_{i+1}}{\sigma_i} = \frac{\sigma_{i+1,1}}{\sigma_{i,1}}
   \; \text{ for } i=1:m-1.\]
\end{definition}
We consider the case $r_i \le r \ll 1$ for $i=1:m-1$, and propose an explicit gradient descent scheme involving multiple learning rates $\eta_i$ for $i=1:m$. The algorithm is defined in~\Cref{alg:mrgd}. 
\begin{algorithm} 
\caption{Multirate Gradient Descent (MrGD)}
\label{alg:mrgd}
\begin{algorithmic}[1]
\STATE {\bfseries Input:}\\ 
    \smallskip
    \quad$\theta^0$: the initial guess (typically zeros or random);\\ 
    \quad$\eta_i$:\!\; the learning rate on the $i$-th scale; \\
    \quad$n_i$: the number of iterations on the $i$-th scale;\\
    \quad$K$: the number of outer iterations.
    \smallskip
\FOR {$k=1:K$}
\STATE \[ \theta_{m,0}^{k} := \theta^{k-1}\]
\FOR {$i=m:-1:1$}
    \FOR {$l=1:n_i$}
    \STATE \begin{equation}\label{eqn:iteration}
    \theta^{k}_{i,l} := \theta^{k}_{i,l-1} - \eta_i \nabla \mathcal L(\theta^k_{i,l-1})
    \end{equation}
    \ENDFOR
\STATE   \[\theta^k_{i-1,0} := \theta^k_{i,n_i}\]
\ENDFOR
\STATE \[\theta^{k} := \theta^{k}_{1,n_1}\]
\ENDFOR
\STATE {\bfseries Output:} $\theta^K$
\end{algorithmic}
\end{algorithm}

To analyze the convergence property of~\Cref{alg:mrgd}, we investigate the error dynamics:

Let $\theta^*$ be the global minimum, which satisfies the optimality condition $A\theta^* = g$, and $e^k_{j,l}:=\theta^k_{j,l}-\theta^*$ be the error at each step where $i$ is replaced by $j$ for clarity in subsequent discussion. Then, based on \Cref{eqn:iteration}, we have $e^k_{j,l} = e^k_{j,l-1} - \eta_jAe^k_{j,l-1}$.
Therefore, the error propagation operator $S$ for one outer iteration can be written as 
\begin{equation}
    S =  Q_1^{n_1} Q_2^{n_2} \cdots   Q_{m-1}^{n_{m-1}} Q_m^{n_m},\, 
\end{equation}
where
\begin{equation}
    Q_j = I - \eta_j \nabla^2 \mathcal L = I - \eta_j A,\; \text{ for } j = 1:m. 
\end{equation}

Given the gaps between different groups of eigenvalues as depicted in~\ref{eqn:spectrum}, for each iteration with $Q_j$, we select learning rates ensuring the reduction of components of $\theta$ in the $j$-th eigenspace. Specifically, we choose
\begin{equation}\label{eqn:learning rate}
    \eta_j\sim\cO\left(\frac{1}{\sigma_j}\right),\;
    \text{ with }  \eta_j\le \frac{1}{\sigma_j},
\end{equation}
or we express it as $\eta_j\lesssim \cO(1/\sigma_j)$. Notably, \Cref{eqn:learning rate} results in a periodic learning rate scheme in~\Cref{alg:mrgd}, with monotonicity maintained within each of the $K$ outer iterations. \Cref{rmk:lr} provides further discussion on $\eta_j$.

The learning rates in~\Cref{eqn:learning rate} also leads to uniform convergence in the $i$-th eigenspace with any $i\ge j$, with the convergence rate being linked to the smallest eigenvalue in the $i$-th group: $\sigma_{i,d_i}$ as outlined in the following remark:

\begin{remark}\label{rmk:convergence}
    For $j\le i$, where $i,j = 1:m$, since $\eta_j\sigma_{i,d_i}\lesssim 
    \sigma_{i,d_i}/\cO(\sigma_j)\le 1$, the convergence rate for $Q_j$, when the parameter $\theta$ is restricted to the eigenspace $V_i$ associated with $\{\sigma_{i,1}, \cdots \sigma_{i,d_i}\}$, is determined by $\eta_j$ and $\sigma_{i, d_i}$, s.t.
    \begin{equation*}
        \sup_{v_{i} \in V_{i},~\|v_{i}\|=1}\|Q_j v_{i}\| \le  (1-\eta_j\sigma_{i,d_i}) =  (1-R^j_i\kappa_i^{-1}),
    \end{equation*}
    where in the last equality, we denote $ R_i^j := \eta_j \sigma_i$ for demonstrating the role of the local condition number $\kappa_i$. For $j>i$, $\sigma_i \gg \sigma_j \implies \eta_j\sigma_i \gg 1$, implying $\theta$ restricted to $V_i$ will be growing, where the growth is related to $\left|1-\eta_j\sigma_i\right|=\left|1-R^j_i\right|\gg 1$.
\end{remark}

\begin{remark}
    For the convenient of further analysis and better readability, we summarize for $R^j_i$ that the relative magnitude of $j$ and $i$ determines if $\theta$ is convergent or growing in the eigenspace $V_i$ under the learning rate $\eta_j$. Specifically, $\theta \text{ restricted to } V_i \text{ is: }$
    \[
    \begin{cases}
        \text{growing with rate } \sim\cO\big(\left|1-R^j_i\right|\big) \gg 1, & \text{if }j>i \text{ in } R^j_i \\
        \text{convergent with rate }\le(1-R^j_i\kappa_i^{-1}), & \text{if }j\le i\text{ in } R^j_i
    \end{cases}\]
\end{remark}


By showing $\|S\| \le 1$, the following theorem indicates~\Cref{alg:mrgd}
is uniformly convergent, provided the inner iteration numbers  $n_1, n_2, \cdots, n_{m-1}$ are chosen appropriately, with $n_m=1$, which will be justified later. 
 
\begin{theorem}\label{thm:Sm} 
    For $n_m=1$ and any $n_{m-1}, n_{m-2}, \cdots, n_1$ satisfying 
    \begin{equation}\label{eq:nicondition}
        n_i \ge \left\lceil \sum_{j=i+1}^m n_{j} F_{i,j}\right\rceil,
    \end{equation}
    where $F_{i, i+1} = $
    {\footnotesize
\begin{equation*}
 \left.  \left(-\log(r_{i}) +\log\left(\frac{\left|r_{i} - R^{i+1}_{i+1}\right|}{1-R^{i+1}_{i+1}\kappa_{i+1}^{-1}}\right) \right)    \right\slash \log\left(\frac{1-R^{i}_{i+1}\kappa_{i+1}^{-1}}{1-R^{i}_{i}\kappa_{i}^{-1}}\right)
\end{equation*}}
\normalsize
and $    F_{i,j} = $
\small
\begin{equation*}
 \left.\left(-\log(r_{i}) + \log\left( \frac{\left|r_i - R_{i+1}^{j}\right|}{\left|1-R_{i+1}^{j}\right|}\right) \right)
        \right \slash \log\left(\frac{1-R^{i}_{i+1}\kappa_{i+1}^{-1}}{1-R^{i}_{i}\kappa_{i}^{-1}}\right)
\end{equation*}
\normalsize
for all $j=i+2:m$ and $R_l^i := \eta_i\sigma_l$, 
    we have
    \begin{equation}
    \|S\| \le \prod_{j=1}^{m} \left(1 - R_m^j\kappa_{m}^{-1}\right)^{n_j}. 
    \end{equation}
\begin{proof}
    See~\Cref{sec:proofSm}
\end{proof}
\end{theorem}

\begin{remark}\label{rmk:n_in_m}
Here, we justify the choice of $n_m=1$. One can rewrite~(\ref{eq:nicondition}) as:  
\begin{align*}
    n_{m-1} &\ge n_mF_{m-1,m},\\
    n_{m-2} &\ge n_mF_{m-2,m} + n_{m-1}F_{m-2,m-1}\\  
    &\ge (F_{m-2,m} + F_{m-1,m}F_{m-2,m-1})n_m,\; \ldots
\end{align*}
indicating the lower bound for any $n_i$ scales linearly with $n_m$, \emph{i.e.}, $n_i \ge c_in_m$ for some constants $c_i$'s. Therefore, for a fixed problem, the final iteration number $n_m$ determines the rest of $n_i$'s.

Furthermore, using the linearity of $Q_i$, the error propagation operator for \Cref{alg:mrgd} can be expressed as:
\begin{align*}
     S^K = \left(Q_1^{n_1} Q_2^{n_2} \cdots   Q_{m-1}^{n_{m-1}} Q_m^{n_m}\right)^K
     = Q_1^{Kn_1} Q_2^{Kn_2} \cdots   Q_{m-1}^{Kn_{m-1}} Q_m^{Kn_m}.
\end{align*}
 
Therefore, for any desired value of $n_m$, one can simply set $n_m=1$ in \Cref{alg:mrgd} and execute the outer iterations the desired number of times.  Thus, we assume $n_m=1$ throughout the remainder.
\end{remark}

\begin{remark}\label{rmk:lr}
The linearity of $Q_i$ also guarantees that the order of iterations associated with $n_i$ does not affect the outcome. This implies that \Cref{alg:mrgd} will converge regardless of whether the learning rate scheme for $\eta_i$'s is monotonic or not, as long as $\eta_i$'s satisfy~\Cref{eqn:learning rate} and the corresponding iteration number $n_i$'s satisfy the result presented in \Cref{thm:Sm}.
\end{remark}

\subsubsection{Comparison with traditional methods}\label{sec:toy_model}
\begin{assumption}\label{asm:kappaeta}
Consider a special case where the local condition numbers and decay rates in the original system~(\ref{eq:A}) are equal and relatively small, respectively, i.e., 
\begin{equation}
\kappa_i  =\kappa_c \gtrsim \cO(1) \quad \text{and} \quad  r_i = r \ll 1.
\end{equation}
The above assumption ensures a 
hierarchical structure in the eigenvalue clusters with a global condition number:
\small
\begin{equation*}
    \kappa = \frac{\sigma_{1,1}}{\sigma_{m,d_m}} = 
    \frac{\sigma_{1,1}}{\sigma_{2,1}} \frac{\sigma_{2,1}}{\sigma_{3,1}} \cdots \frac{\sigma_{m-1,1}}{\sigma_{m,1}} \frac{\sigma_{m,1}}{\sigma_{m,d_m}} = \kappa_c r^{1-m} \gg 1. 
\end{equation*}
\normalsize
In addition, following~\Cref{eqn:learning rate}, we set the multirate $\eta_i$'s based on some constant $\eta \gtrapprox 1$, \emph{s.t.},
\begin{equation*}
    \eta_i = \frac{1}{\eta\sigma_i} \lesssim \cO\left(\frac{1}{\sigma_i}\right).
\end{equation*}
\end{assumption}

Thus, for the vanilla GD with a constant learning rate $\eta_1$, we need $\eta_1 \lesssim \cO(\frac{1}{\sigma_1})$ to ensure convergence and the convergence rate is then given by $\big(1-1/(\kappa_c r^{1-m})\big)$. To achieve an error of $\cO(\varepsilon)$:
\begin{equation}\label{eqn:conv_step}
    \left( 1-\frac{1}{\kappa_cr^{1-m}}\right)^n <\cO(\varepsilon) \implies n \gtrsim \mathcal O \left( \kappa_c r^{1-m} |\log(\varepsilon)|\right)
\end{equation}
we need $n$ many vanilla GD steps as indicated above.
For accelerating methods, such as Conjugate Gradient~\cite{hestenes1952methods}, GD with momentum~\cite{polyak1964some}, Chebyshev iterations~\cite{manteuffel1977tchebychev}, Nestrov acceleration~\cite{nesterov1983method}, etc., one may need
\begin{equation}\label{eqn:gd+}
    \mathcal O \left( \sqrt{\kappa_c r^{1-m}} |\log(\varepsilon)|\right).
\end{equation}

Now, we demonstrate how MrGD can accelerate convergence by using multiple  learning rates. 
As a direct consequence of Theorem~\ref{thm:Sm}, we have the following uniform convergence result:
\begin{corollary}\label{cor:SmSimple}
Under Assumption~\ref{asm:kappaeta}, \Cref{thm:Sm} yields:
\begin{equation}
    \|S\| \le \prod_{j=1}^{m} \left(1 - R_m^j\kappa_{m}^{-1}\right)^{n_j} 
    \le \left(1-\frac{1}{\eta\kappa_c}\right). 
\end{equation} 
\end{corollary}
Since $\kappa_c\gtrsim \cO(1) $, we can also choose a constant $\eta>1$ such that $ \frac{1}{\eta\kappa_c}\lesssim\cO(1)$. Following the same derivation as in~(\ref{eqn:conv_step}), we need only $\cO(|\log(\varepsilon)|)$ outer iterations in MrGD to achieve an error of $\cO(\varepsilon)$. The number of total GD iterations with different learning rates is then:
\begin{equation}
    \mathcal O\left(n |\log(\varepsilon)|\right),
\end{equation}
where $n=\sum_{i=1}^m n_i$ is the number of GD iterations in one step of outer iteration.

The only question left is how $n$ compares with $\kappa_c r^{1-m}$ under Assumption~\ref{asm:kappaeta}.
As mentioned in~\Cref{rmk:n_in_m}, the lower bound for $n$ is characterized completely by $F_{i,j}$ for any $j\ge i+1$. As a result, we provide upper bounds for $F_{i,j}$ below under a more general case than Assumption~\ref{asm:kappaeta}:

\begin{corollary}\label{cor:FijGr}
    For any fixed $0<r\ll 1$ and  $\eta\kappa_c\gtrsim \cO(1)$, we have $F_{i,i+1},\,F_{i,j} \ge 0$, and
    \small
     \[  \begin{cases}
     F_{i,i+1}\le \displaystyle\frac{\eta\kappa_c(\eta\kappa_c - r)\big.\left(-\log\big(r(\eta\kappa_c - 1)\big) + \kappa_c -\eta r -1\big.\right)}{\eta\kappa_c(1-r)-r}
   \\ &\\
   F_{i,j}\le \displaystyle\frac{\eta\kappa_c(\eta\kappa_c - r)\left(-\log(r) + C r^{j-(i+1)}\right)}{\eta\kappa_c(1-r)-r}, \quad 
    \end{cases}
    \]
    \normalsize
    \text{ for $j=i+2:m$.} This means, as $r \to 0$, we have
    \begin{equation}
        F_{i,j} = \mathcal O\left(\eta\kappa_c \left|\log(r)\right|\right) 
    \end{equation}
    for all $j=i+1:m$, which leads to the estimate for $n$:
    \begin{equation}
        n = \mathcal O \left( (\eta\kappa_c)^{m-1} \left|\log(r)\right|^{m-1}\right).
    \end{equation}
\begin{proof}
    See~\Cref{proof:FijGr}.
\end{proof}
\end{corollary}



The scenario described addresses problems possessing pronounced hierarchical spectral structures in the asymptotic regime as $r\to 0$. For these problems, MrGD demonstrates significant enhancements, attaining an improvement factor of $|\log(r)|$, in contrast to the polynomial improvement factor seen within other gradient descent techniques.

Lastly, we examine a harmless scenario where $\kappa_c\gtrapprox 1$ as a basic validation step. In this instance, having knowledge of the eigenvalue groups essentially equates to knowing almost all the eigenvalues. Any algorithm utilizing such information should ideally demonstrate rapid convergence.
Indeed, for MrGD, for any fixed $r > 0$ and $\kappa_c \gtrapprox 1$, as $\eta \to 1^+$ such that $\eta\kappa_c \to 1^+$, we have
\begin{equation}
F_{i,i+1} \to 1 \text{ and } F_{i,j} \to 0, \text{ for } j = i+2:m.
\end{equation}
This results in $n_j \approx n_m = 1$ for all $j = 1:m-1$, leading to  $n = \sum_{i=1}^m n_i \to m$, and an overall complexity of $\mathcal{O}(m|\log(\varepsilon)|)$ for MrGD. This aligns with our intuition as the MrGD method effectively adopts the learning rate $\frac{1}{\sigma_i}$ for the iterations associated with $n_i = 1$, resulting in a one-step convergence for group $i$. 

Furthermore, for certain specific problems, the multigrid methods~\cite{xu1992iterative,hackbusch2013multi} 
also achieve a similar rapid convergence of complexity $\mathcal{O}(m|\log(\varepsilon)|)$.  However, as highlighted in~\cite{lee2007robust,xu2017algebraic}, constructing such an efficient multigrid method typically requires detailed knowledge of the operator's eigenspace.

\begin{figure}[ht]
\vskip 0.2in
\begin{center}
\hspace*{5em}
\begin{subfigure}[b]{0.32\linewidth}
            \centering
            \includegraphics[width=\textwidth]{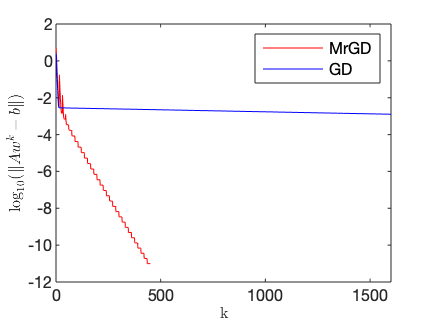}
            \caption[]%
            {Two scales with $r=0.001$}    
        \end{subfigure}
        \hspace*{\fill}
        \begin{subfigure}[b]{0.32\linewidth}  
            \centering 
            \includegraphics[width=\textwidth]{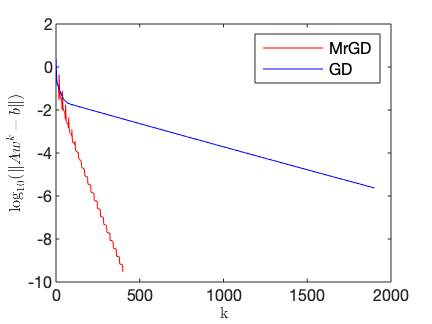}
            \caption[]%
            {Three scales with $r=0.1$}   
        \end{subfigure}
        \hspace*{\fill}
\caption{Numerical examples for randomly generated 100-dimensional linear regression problem with two scales ($m=1$, $r = 0.001$) and three scales ($m=2$, $r=0.1$). More details of this example can be found in Appendix~\ref{appendix_linearexample}.}
\label{figs:MrGDlinear}
\end{center}
\vskip -0.2in
\end{figure}

In Table~\ref{tab:comp}, the methods discussed above are compared in terms of computational complexity and necessary information. Figure~\ref{figs:MrGDlinear} illustrates two examples of convergence profiles for GD and MrGD on 100-dimensional linear regression problems with different number of scales.

\begin{table}[h]
\caption{Comparison of time complexities required to achieve $\cO(\varepsilon)$-error using various methods, along with the corresponding information needed. ``e.val" stands for eigenvalues, and ``GD+" stands for gradient decent methods with accelerations related to~(\ref{eqn:gd+}). $\kappa_c \approx 1$ is the local condition number, and $r\ll 1$ is the decay rate of eigenvalues.} 
\label{tab:comp}
\begin{center}
\small
\begin{tabular}{c cccc}
\multicolumn{1}{c}{\bf  }  &\multicolumn{1}{c}{GD} &\multicolumn{1}{c}{GD+} &\multicolumn{1}{c}{MrGD} &\multicolumn{1}{c}{Multigrid}
\\ \hline \\
\bf Complexity       & $\mathcal O\left(\displaystyle\frac{\kappa_c |\log(\varepsilon)|}{r^{m-1}}\right)$ & $\mathcal O\left(\displaystyle\frac{\sqrt{\kappa_c} |\log(\varepsilon)|}{\sqrt{r^{m-1}}}\right)$&  $\mathcal O\left(\left(\eta\kappa_c|\log(r)|\right)^{m-1}\left|\log(\varepsilon)\right|\right)$ & $\mathcal O(m|\log(\varepsilon)|)$ \smallskip\\
\hline\\
\bf Information &largest e.val  & largest e.val & e.val clusters & eigenspace \smallskip\\
\hline
\end{tabular}
\end{center}
\end{table}

\subsection{Multirate gradient descent for convex problems}
In this section, we extend the insights obtained from the quadratic minimization problems~(\ref{eq:A}) to more general convex minimization problems. Consider
\begin{equation*}
    \min_{\theta \in \mathbb R^d} \mathcal L(\theta),
\end{equation*}
for $\mathcal L:\mathbb R^d \mapsto \mathbb R$ $L$-smooth and $\mu$-strongly convex, $i.e.$,
\begin{equation}\label{eq:convexP}
     \mu\mathbf I_d  \preceq \nabla^2 \mathcal L (\theta) \preceq L \mathbf I_d.
\end{equation}
We introduce the following assumptions: 
\begin{assumption}\label{asm:ms-spectrum}
There exists an orthogonal matrix $\Pi\in\bbR^{d\times d}$ of the form:
\begin{equation}\label{eqn:Pi}
    \Pi^T = \begin{bmatrix}
    \Pi_1^T | \Pi_2^T | \dots |\Pi_m^T
\end{bmatrix}\in\bbR^{d\times d},
\end{equation}
with $\Pi_i\in\bbR^{d_i\times d}$ such that
\begin{equation}
    \sigma_{i,d_i} \mathbf I_{d_i} \preceq \Pi_i \nabla^2  \mathcal L (\theta)  \Pi_i^T\preceq \sigma_{i,1} \mathbf I_{d_i}.
\end{equation}
for $\sigma_{i,1},\,\sigma_{i,d_i}$ satisfying~(\ref{eqn:spectrum}) for $i=1:m$.
Furthermore, we assume the cross-spectrum is bounded by a small positive number $0 \le \delta \ll 1$ for any $i\neq j$:
\begin{equation}\label{eqn:c-spectrum}
    \left\|  \Pi_i \nabla^2  \mathcal L (\theta)  \Pi_j^T\right\| \le \delta.
\end{equation}
\end{assumption}
Then, the convergence result of the MrGD~\Cref{alg:mrgd} introduced in~\Cref{thm:Sm} can be generalized to convex problems by the following theorem:
\begin{theorem}\label{thm:ConvexConvergence}
 Let $\theta^k_{i,l}$ be computed by~\Cref{alg:mrgd} for minimizing~(\ref{eq:convexP}), with
 iteration numbers $n_{m-1}, n_{m-2}, \cdots, n_1$ satisfying~(\ref{eq:nicondition}). 
  If $\|\theta^k_{i,l} - \theta^*\| \le C$ for all $i=1:m$ and $l=1:n_i$, for some constant C, under Assumptions~\ref{asm:ms-spectrum},
    \begin{align*}
        \|\theta^k - \theta^*\| \le \prod_{j=1}^{m} \left(1 - R_m^j\kappa_{m}^{-1}\right)^{n_j}\|\theta^{k-1} - \theta^*\|
        + \delta m C \max_{i=1:m} \left\{ \sum_{j=0}^{m-1}\eta_{j+1}C_i^jE_i^j\right\},
    \end{align*}
    where, when $j \le i$, 
    \[
        C_i^j = \prod_{s=1}^j (1-R_i^s\kappa_i^{-1})^{n_{s}} \quad \text{and}\quad E_i^j = \sum_{s=0}^{n_j-1} \left(1-R_i^j\kappa_i^{-1}\right)^s,
    \]
    and when $j>i$,
    \[C_i^j = \prod_{s=1}^{i} \left(1-R_i^s\kappa_i^{-1}\right)^{n_{s}} \prod_{s=i+1}^{j} \left|1-R_i^s\right|^{n_{s}} \quad \text{and}\quad E_i^j = \sum_{s=0}^{n_j-1} \left|1-R_i^j\right|^s,\]
    and $R_i^j = \eta_j \sigma_{i,1} =\eta_j\sigma_i$ as defined in~\Cref{rmk:convergence}.
    
\begin{proof}
    See~\Cref{sec:proof_thm4}.
\end{proof}
\end{theorem}
The following Theorem states that 
if the cross-spectrum $\delta$ vanishes, we can recover results identical to~\Cref{thm:Sm}:
\begin{corollary}\label{thm:Convex0}
With assumption~(\ref{eqn:c-spectrum}) replaced by $\left\|  \Pi_i \nabla^2  \mathcal L (\theta)  \Pi_j^T\right\| =0$ for any $i\neq j$, and all other hypotheses in ~\Cref{thm:ConvexConvergence} standing,  
    \begin{equation}
        \|\theta^k - \theta^*\| \le \prod_{j=1}^{m} \left(1 - R_m^j\kappa_{m}^{-1}\right)^{n_j}\|\theta^{k-1} - \theta^*\|.
    \end{equation}
Furthermore, under Assumption~\ref{asm:kappaeta},  we have 
    \begin{equation}
        \|\theta^{k} - \theta^*\| \le  \left(1 - \frac{1}{\eta\kappa_c}\right)\|\theta^{k-1} - \theta^*\|.
    \end{equation}
\begin{proof}
    Take $\delta=0$ in~\Cref{thm:ConvexConvergence}
\end{proof}
\end{corollary}
An illustrative example of the above scenario is when the loss function takes the form:
\begin{equation}\label{eq:nonquadraticLoss}
    \mathcal L(\theta) = \frac{1}{2}\theta^T A \theta - g^T\theta + \sum_{i=1}^m f_i(\Pi_i \theta),
\end{equation}
where $\Pi = \left[\Pi_1^T | \Pi_2^T | \dots |\Pi_m^T \right]\in\bbR^{d\times d}$ forms the spectral decomposition of $A$ with $\frac{1}{2}\sigma_{i,d_i} \mathbf I_{d_i}\preceq \Pi_i A \Pi^T_i \preceq \frac{1}{2}\sigma_{i,1} \mathbf I_{d_i}$ and $f_i: \mathbb R^{d_i} \mapsto \mathbb R$ are arbitrary convex functions with $\frac{1}{2}\sigma_{i,d_i} \mathbf I_{d_i}\preceq \nabla^2 f_i(\theta) \preceq \frac{1}{2}\sigma_{i,1} \mathbf I_{d_i}$ for all $i=1:m$ and $\theta$. Then, by direct calculations, we have 
$$
\begin{aligned}
\nabla^2 \mathcal L(\theta) = A + \Pi {\rm diag}\left(\nabla^2f_1, \nabla^2f_2, \cdots, \nabla^2f_m \right)\Pi^T 
=\Pi{\rm diag}(
 \Pi_1 A \Pi^T_1 + \nabla^2f_1,\cdots, \Pi_m A \Pi^T_m + \nabla^2f_m)\Pi.
\end{aligned}
$$
That is, under the above conditions, \Cref{eq:nonquadraticLoss} provides a general example of a non-quadratic loss satisfying $\delta=\Pi_i \nabla^2 \mathcal L(\theta) \Pi^T_j =0$ for all $i\neq j$.  

In addition, the small cross-spectrum is pertinent in the case of local convergence. Specifically, we express $\nabla^2 \mathcal L(\theta) = \nabla^2 \mathcal L(\theta^*) + \left(\nabla^2 \mathcal L(\theta) - \nabla^2 \mathcal L(\theta^*)\right)$ and take $\Pi$ to be the eigenvectors of $\nabla^2 \cL(\theta^*)$, then the cross-spectrum is given by $\left\|\Pi_i \left(\nabla^2 \mathcal L(\theta) - \nabla^2 \mathcal L(\theta^*)\right)\Pi_j^T\right\|$, which can be sufficiently small if $\theta$ is very close to $\theta^*$, the global minimizer of the loss $\mathcal L(\theta)$.

\section{Conclusion}
This paper shows that multiscale data leads to empirical loss functions whose gradients and Hessians inherit the multiscale properties. Utilizing these properties, the introduced explicit MrGD scheme converges with a near-optimal rate for convex problems. The theories developed in this work partially explain the so-called learning rate warm-up strategy employed in neural network training.

 \section*{Acknowledgements}

Liu and Tsai are partially supported by National Science Foundation Grant DMS-2110895. Tsai is also supported partially by Army Research Office Grant W911NF2320240. He is supported by the KAUST Baseline Research Fund.


\bibliography{ICML2024_bib}
\bibliographystyle{icml2024}

\newpage
\appendix
\onecolumn
\section{Notations for the proofs}
Here we list out some notations we used for the proofs in the appendices.
\begin{notation}\label{notation:1}
	Let $z$ be a function of $({\vec  x},g)$ or a random variable in $\mathbb{R}^m$ or $\mathbb R^{m\times n}$ over some probability space and let $z_i$ denote a sample realization of $z$. We denote the empirical average on the whole dataset
	\begin{equation*}
		\left< z\right>_N :=  \frac{1}{N}\sum_{i=1}^N z(({\vec  x_i},g)),
	\end{equation*}
 the average on the mini-batch $\mathcal B \subset \mathcal D$
 \begin{equation*}
		\left< z\right>_{\mathcal B} := \frac{1}{|\mathcal B|}\sum_{{({\vec  x},g) \in B}} z(({\vec  x},g)),
\end{equation*}
	and the mean
	\begin{equation*}
		\left<z \right> :=\lim_{N\rightarrow\infty} \left< z\right>_N = \mathbb{E}[z].
	\end{equation*}
\end{notation}

\begin{notation}
Here, we introduce some commonly used products for tensors to simplify our notation. For simplicity, we only introduce these products under special cases, which will appear in the following sections. For any r-th order tensor $A \in \mathbb R^{n_1 \times n_2 \times \cdots n_r}$, let us use 
\begin{equation}
    \left[ A \right]_{i_i,i_2, \cdots, i_r}, \quad i_1=1:n_1, i_2=1:n_2, \cdots, i_r=1:n_r
\end{equation}
to define elements in $A$ with index $(i_1, i_2, \cdots, i_r)$ for $i_1=1:n_1, i_2=1:n_2, \cdots, i_r=1:n_r$.

    \begin{description}
    \item[Tensor contraction:] Let $A\in \mathbb R^{n_1\times n_2\times n_r\times k }$ and $B\in \mathbb R^{k\times m_1\times m_2\times m_s}$, we denote the tensor contraction $A \cdot B \in \mathbb R^{n_1\times \cdots \times n_r\times m_1\times \cdots\times  m_s}$ as
    \begin{equation}\label{eq:TC}
        \left[A \cdot B\right]_{i_1,\cdots, i_r,j_1,\cdots,j_s} = \sum_{t=1}^k \left[A\right]_{i_1,\cdots,i_r,t}\left[B\right]_{t,j_1,\cdots, j_s}
    \end{equation}
    which then becomes a 2nd order tensor. 
    \item[Tenor production:] Let $A\in \mathbb R^{n\times m}$ and $b\in \mathbb R^{k}$, we denote the tensor production $A\otimes b \in \mathbb R^{n\times m\times k}$ as
    \begin{equation}\label{eq:TP}
        \left[A\otimes b\right]_{i,j,k} = \left[A\right]_{i,j}\left[b\right]_s, \quad i=1:n,~ j=1:m, ~ s=1:k,
    \end{equation}
    which then becomes a 3rd order tensor.
    \item[Hadamard product:] Let $a, b\in \mathbb R^{k}$, we denote the Hadamard product $a\odot b\in  \mathbb R^{k}$ as
    \begin{equation}\label{eq:HP}
        \left[a \odot b\right]_{s} = \left[a\right]_{s}\left[b\right]_s, \quad s=1:k,
    \end{equation}
    which is still a 1st order tensor.
    \end{description}
\end{notation}
\begin{notation}
    Here we denote some norms for vector and matrix spaces. 
    For vectors, we denote the $\ell^p$ norm of vector $a \in \mathbb R^d$ as
    \begin{equation}
        \|a\|_{p} := \left( \sum_{i=1}^d |a_i|^p\right)^{\frac{1}{p}}
    \end{equation}
    for all $1\le p\le \infty$. In particular, we have
    \begin{equation}
        \|a\|_{\infty} := \max_{i} \left\{ |a_i| \right\}.
    \end{equation}
    For simplicity, we define $\|a\| := \|a\|_{2}$ for vectors.
    
    For matrix, we introduce only the $\ell^2$ norm for any $A \in \mathbb R^{n\times m}$ as
    \begin{equation}
        \|A\| = \sup_{b \in \mathbb R^m, \|b\| \le 1} \frac{\|Ab\|}{\|b\|}.
    \end{equation}
\end{notation}

\begin{lemma}\label{lemm:vectornorm}
    For any $a \in \mathbb R^{d}$, we have
\begin{equation}
    \|a\|_{\infty} \le \|a\| \le \|a\|_{1} \le n\|a\|_\infty.
\end{equation}
\end{lemma}

\begin{remark}
For any $a \in \mathbb R^{1\times d}$, which can be understood as either a matrix or a vector, the $\ell^2$ matrix norm is consistent with the $\ell^2$ vector norm. By abuse of notation, we will always use $\|\cdot\|$ for $\ell^2$ norm for both vectors and matrices.
\end{remark}

\section{Multiscale structure of gradient flow in DNNs}\label{sec:proof_dnn_expansion}
In this section, we showcase that there exists the same structure in the gradient flow of DNNs compared to linear and logistic regressions when data have the same multiscale structure. For simplicity, let us first prove the result for data with two scales. Then, we generalize our results to multiscale data by induction.
\subsection{Two-scale structure of gradient flow in DNNs}
For data with two scales, we assume the data points concentrate around a low-dimensional subspace with a small variance in the orthogonal complement. More specifically, 
\begin{equation}\label{eq:Dtwoscale}
    {\vec x_i} = \begin{pmatrix}
    x^0_i \\ \varepsilon x^1_i
    \end{pmatrix} \in \mathbb R^{d_0 + d_1},
\end{equation}
where $x^0_i \sim \mathcal U_x \in \mathbb R^{d_0}$, $x^1_i \sim \mathcal U\left([-1,1]^{d_1}\right) \in \mathbb R^{d_1}$, and $0 \le \varepsilon \ll 1$. Following the notation above, we have
\begin{equation}
    \vec x^0 = \begin{pmatrix}
        x^0 \\ 0
    \end{pmatrix}, \quad \vec x^1 = \begin{pmatrix}
        x^0 \\ \varepsilon x^1
    \end{pmatrix} = \vec x.
\end{equation}

In the rest of this subsection, we prove the two-scale structure of the gradient flow for DNNs. First, have
\begin{equation}
    \frac{\partial \mathcal L}{\partial W^{\ell}} = \left< (f(\vec x;\theta)-g) \frac{\partial f}{\partial f^{\ell}} \cdot \frac{\partial f^{\ell}}{\partial W^{\ell}} \right>_N,
\end{equation}
where $\frac{\partial f}{\partial f^{\ell}} \cdot \frac{\partial f^{\ell}}{\partial W^{\ell}} \in \mathbb R^{n_{\ell} \times n_{\ell-1}}$ denotes the tensor contraction between the 1st order tensor $\frac{\partial f}{\partial f^{\ell}}\in \mathbb R^{n_\ell}$ and the 3rd order tensor $\frac{\partial f^{\ell}}{\partial W^{\ell}} \in \mathbb R^{n_\ell \times n_\ell \times n_{\ell-1}}$ as in~~\Cref{eq:TC}. 

For $\frac{\partial f^{\ell}}{\partial W^{\ell}}$, we have
\begin{equation}
    \begin{aligned}
     \frac{\partial f^{\ell}(\vec x)}{\partial W^{\ell}} = {\rm diag} \left(D^\ell(\vec x)\right) \otimes f^{l-1}(\vec x),
    \end{aligned}
\end{equation}
where 
\begin{equation}
    D^\ell(\vec x) =  \sigma'\left( W^\ell f^{\ell-1}(\vec x) + b^\ell\right)  \in \mathbb R^{n_\ell}
\end{equation}
and $\otimes$ denotes the tensor product as in~~\Cref{eq:TP}. Furthermore, let us denote 
\begin{equation}
    \lambda^\ell(\vec x) = \left. \frac{\partial f}{\partial f^{\ell}} \right|_{\vec x}.
\end{equation}
Given the properties of the tensor product and standard matrix/vector product, we have
\begin{equation}
    \frac{\partial f}{\partial f^{\ell}} \cdot \frac{\partial f^{\ell}}{\partial W^{\ell}} = \lambda^\ell(\vec x) \cdot {\rm diag}\left(D^\ell(\vec x)\right) \otimes f^{\ell-1}(\vec x) = \left( \lambda^\ell(\vec x) \odot D^\ell(\vec x) \right) \left(f^{\ell-1}(\vec x)\right)^T,
\end{equation}
where $\odot$ denotes the Hadamard product as in~~\Cref{eq:HP}. Here, $\lambda^\ell(\vec x) \odot D^\ell(\vec x)$ will be the element-wise product with two vectors in $\mathbb R^{n_\ell}$. We further denote 
\begin{equation}\label{eqn:lambda_tilde}
    \widetilde \lambda^\ell(\vec x) = \lambda^\ell(\vec x) \odot D^\ell(\vec x) \in \mathbb R^{n_\ell}.
\end{equation}

Consequently, we can rewrite $\frac{\partial \mathcal L}{\partial W^{\ell}}$ as
\begin{equation}
    \frac{\partial \mathcal L}{\partial W^{\ell}} = \left< (f(\vec x;\theta)-g)  \widetilde \lambda^\ell(\vec x) \left(f^{\ell-1}(\vec x)\right)^T \right>_N
\end{equation}

\subsubsection{Two-scale decomposition of $f^\ell$}
We first show the two-scale structures in the forward propagation process in DNNs with general activation functions.
\begin{lemma}\label{lemm:fl}
For any DNN defined in~\Cref{eq:defdnn} and dataset with two scales as in~\Cref{eq:Dtwoscale} under Assumption~\ref{assumption:1}, we have
\begin{equation}
    f^{\ell}(\vec x) = f^\ell_0(\vec x^0) + \varepsilon f^{\ell}_1(\vec x^1), \quad \ell=1:L
\end{equation}
where 
\begin{equation}
    f^\ell_0(\vec x^0) = f^\ell(\vec x^0) = f^\ell\left( \begin{pmatrix}
        x \\0
    \end{pmatrix}\right)
\end{equation}
and 
\begin{equation}
    \|f^{\ell}_1(\vec x^1)\| \le \left({\rm Lip}(\sigma)\right)^{\ell} \left\|W^1_1 x^1 \right\| \prod_{k=2}^\ell \left\|W^k\right\| ,
\end{equation}
where ${\rm Lip}(\sigma)$ denotes the Lipschitz constant of activation function $\sigma$.
\end{lemma}

\begin{proof}
We use proof by induction. 
For $\ell=1$, we have
    \begin{equation}
        f^{1}(\vec x) = \sigma\left( W^1_0 x^0 + b^1 + \varepsilon W^1_1x^1\right),
    \end{equation}
where $W^1 = \begin{pmatrix}
    W^1_0 \\ W^1_1
\end{pmatrix}$ and $W^1_0, W^1_1$ correspond to the $x^0, x^1$ components of $\vec x$ respectively.
We first define 
\begin{equation}
    f^1_0(\vec x^0) = \sigma\left(W^1_0 x^0 + b^1\right) = f^1(\vec x^0) = f^1\left(\begin{pmatrix}
        x \\ 0 
    \end{pmatrix}\right).
\end{equation}
Then, we have 
\begin{equation}
    \begin{aligned}
   \left[f^{1}(\vec x) - f^1_0(\vec x^0)\right]_i &=  \left[\sigma\left( W^1_0 x^0 + b^1 + \varepsilon W^1_1x^1\right) - \sigma\left( W^1_0 x^0 + b^1 \right) \right]_i \\
        &= \varepsilon \sigma'(\xi_{\left[W^1 \vec x + b^1\right]_i, \left[ W^1_1 \varepsilon x^1 \right]_i})  \left[ W^1_1 x^1 \right]_i,
    \end{aligned}
\end{equation}
where $\xi_{\left[W^1 \vec x + b^1\right]_i, \left[ W^1_1 \varepsilon x^1 \right]_i}$ depends on $\left[W^1 \vec x + b^1\right]_i$, and $\left[ W^1_1 \varepsilon x^1 \right]_i$. Since $\varepsilon x^1 $ is a part of $\vec x^1 = \begin{pmatrix}
    x^0 \\ \varepsilon x^1
\end{pmatrix} = \vec x$, it means that $\xi_{\left[W^1 \vec x + b^1\right]_i, \left[ W^1_1 \varepsilon x^1 \right]_i}$ depends on $\vec x^1$ with parameters $W^1$ and $b^1$ which are parts of $\theta$. Thus, we can write 
\begin{equation}
    f^{1}(\vec x) - f^1_0(\vec x^0) = \varepsilon f^1_1(\vec x^1; \theta ) = \varepsilon f^1_1(\vec x^1).
\end{equation}
Moreover, we have
\begin{equation}
    \|\varepsilon f^1_1(\vec x^1)\| = \left\| \sigma\left( W^1_0 x^0 + b^1 + \varepsilon W^1_1x^1\right) - \sigma\left( W^1_0 x^0 + b^1 \right) \right\| \le \varepsilon {\rm Lip}(\sigma)\|W^1_1 x^1\|.
\end{equation}
This finishes the proof for $\ell=1$. \emph{Note that $f^1_1(\vec x^1)$ depends also on the previous scale such that $f^1_1(\vec x^1)=f^1_1(\vec x_0, \vec x_1)$, but for brevity we only include the component corresponding to the smallest scale. The same convention will be adopted throughout the rest of the proof.}

We prove them by induction for $\ell>1$. Now, we assume that results hold for $1, 2, \cdots, \ell-1$, let us prove it for $\ell$. According to the definition and assumption for $\ell-1$, we have
\begin{equation}
f^{\ell}(\vec x) = \sigma\left( W^\ell f^{\ell-1}(\vec x) + b^\ell \right) = \sigma\left( W^\ell (f^{\ell-1}_0(\vec x^0) + \varepsilon f^{\ell-1}_1(\vec x^1)) + b^\ell \right).
\end{equation}
Similarly, we have
\begin{equation}
    f^\ell_0(\vec x^0) := \sigma\left(W^\ell f^{\ell-1}_0(\vec x^0) 
 + b^\ell \right) = \sigma\left(W^\ell f^{\ell-1}\left(\vec x^0\right) + b^\ell \right)
 = f^{\ell}\left(\vec x^0\right).
\end{equation}
Again,
\begin{equation}
    \begin{aligned}
   \left[f^{\ell}(\vec x) - f^\ell_0(\vec x^0)\right]_i &=  \left[\sigma\left( W^\ell f^{\ell-1}_0(\vec x^0) + b^1 + \varepsilon f^{\ell-1}_1(\vec x^1)\right) - \sigma\left( W^\ell f^{\ell-1}_0(\vec x^0) + b^1 \right) \right]_i \\
        &= \varepsilon \sigma'(\xi_{\left[W^\ell f^{\ell-1}_0(\vec x^0) + b^\ell\right]_i, \left[ \varepsilon f^{\ell-1}_1(\vec x^1) \right]_i})  \left[ W^1_1 x^1 \right]_i \\
        &=\varepsilon f^\ell_1(\vec x^1;\theta) = \varepsilon f^\ell_1(\vec x^1). 
    \end{aligned}
\end{equation}
Furthermore, we have
 \begin{equation}
 \begin{aligned}
       \left\| \varepsilon f^{\ell}_1\right\| 
       &=  \left\|\sigma\left( W^\ell (f^{\ell-1}_0(\vec x^0) + \varepsilon f^{\ell-1}_1(\vec x^1)) + b^\ell \right) -  \sigma\left( W^\ell f^{\ell-1}_0(\vec x^0)  + b^\ell \right)
        \right\| \\
        &\le \varepsilon {\rm Lip}(\sigma) \left\| W^\ell f^{\ell-1}_1(\vec x^1)) \right\| \\
        &\le \varepsilon {\rm Lip}(\sigma) \left\| W^\ell \right\| \left\| f^{\ell-1}_1(\vec x^1)) \right\|.
 \end{aligned}
\end{equation}
Finally, we have the bound for $\|f^{\ell}_1(\vec x^1)\|$ by induction. 
\end{proof}

\subsubsection{Two-scale structure in $\widetilde \lambda^\ell$.}
Recall that $\frac{\partial \mathcal L}{\partial W^{\ell}} = \left< (f(\vec x;\theta)-g)  \widetilde \lambda^\ell(\vec x) \left(f^{\ell-1}(\vec x)\right)^T \right>_N$, we will study the two-scale structure in $\lambda^\ell(\vec x)$ and $\widetilde \lambda^\ell$ in this subsection.
Before that, we first have the following two-scale structure in $D^\ell(\vec x):= \sigma'\left( W^\ell f^{\ell-1}(\vec x) + b^\ell\right)  \in \mathbb R^{n_\ell}$.
\begin{lemma}\label{lemm:Dl}
For any DNN defined in~\Cref{eq:defdnn} and dataset with two scales as in~\Cref{eq:Dtwoscale} under Assumption~\ref{assumption:1}, we have 
\begin{equation}
    D^\ell(\vec x)  = D^{\ell}_0(\vec x^0) + \varepsilon D^{\ell}_1(\vec x^1), \quad \ell=1:L.
\end{equation}
In particular, we have
\begin{equation}
    D^{\ell}_0(\vec x^0) = \sigma'\left( W^\ell f^{\ell-1}_0(\vec x^0) + b^\ell\right) = \sigma'\left( W^\ell f^{\ell-1}\left(\vec x^0\right) + b^\ell\right) = D^\ell (\vec x^0),
\end{equation}
and 
\begin{equation}
    \|D^\ell_1(\vec x^1)\| \le {\rm Lip}(\sigma') \left({\rm Lip}(\sigma)\right)^{\ell-1} \left\|W^1_1 x^1 \right\| \prod_{k=2}^\ell \left\|W^k\right\|.
\end{equation}
\end{lemma}
\begin{proof}
Given $\sigma \in C^2$ and the decomposition of $f^{\ell}(\vec x)$ in Lemma~\ref{lemm:fl}, we have
\begin{equation}
    \begin{aligned}
        D^\ell(\vec x) &= \sigma'\left( W^\ell f^{\ell-1}(\vec x) + b^\ell\right) \\
        &=\sigma'\left(W^\ell f^{\ell-1}_0(\vec x^0) + b^\ell + \varepsilon W^\ell f^{\ell-1}_1(\vec x^1)\right) \\
        &= \sigma'\left(W^\ell f^{\ell-1}(\vec x^0) + b^\ell\right) + \varepsilon{\rm diag}\left(\sigma''(\xi)\right)W^\ell f^{\ell-1}_1(\vec x^1)\\
        &=D^\ell_0(\vec x^0) + \varepsilon D^\ell_1(\vec x^1;\theta) \\
        &=D^\ell(\vec x^0) + \varepsilon D^\ell_1(\vec x^1),
    \end{aligned}
\end{equation}
where $\xi \in \mathbb R^{n_\ell}$ and each $\left[\xi\right]_i$ is determined by $\left[W^\ell f^{\ell-1}_0(\vec x^0) + b^\ell \right]_i$, and $\varepsilon \left[W^\ell f^{\ell-1}_1(\vec x^1)\right]_i$ via the intermediate value theorem.
In addition, we have the following bound
    \begin{equation}
    \begin{aligned}
    \left\|\varepsilon D^{\ell}_1 \right\| &= \left\| \sigma'\left(W^\ell f^{\ell-1}_0(\vec x^0) + b^\ell + \varepsilon W^\ell f^{\ell-1}_1(\vec x^1)\right) - \sigma'\left(W^\ell f^{\ell-1}_0(\vec x^0) + b^\ell\right)\right\| \\
    &\le \varepsilon {\rm Lip}(\sigma') \left\| W^\ell f^{\ell-1}_1(\vec x^1)\right\| \\
    &\le\varepsilon {\rm Lip}(\sigma') \left({\rm Lip}(\sigma)\right)^{\ell-1} \left\|W^1_1 x^1 \right\| \prod_{k=2}^\ell \left\|W^k\right\|.
    \end{aligned}
\end{equation}
\end{proof}

\begin{corollary}
We have
\begin{equation}
    \left\|{\rm diag}\left(D^{\ell}_1\right) \right\| \le \|D^\ell_1\|.
\end{equation}
\end{corollary}
\begin{proof} Given the properties of matrix norm and vector norm, we have
    \begin{equation}
         \left\|{\rm diag}\left(D^{\ell}_1\right) \right\|  = \|D^\ell_1\|_\infty \le \|D^\ell_1\|.
    \end{equation}
\end{proof}

Based on the two-scale structure in $D^\ell(\vec x)$, we have the following two-scale structure for $\lambda^\ell(\vec x)$.
\begin{lemma}\label{lemm:ts-lambdal}
For any DNN defined in~\Cref{eq:defdnn} and dataset with two scales as in~\Cref{eq:Dtwoscale} under Assumption~\ref{assumption:1}, we have 
\begin{equation}\label{eq:ts-lambda}
    \lambda^\ell(\vec x)  = \lambda^{\ell}_0(\vec x^0) + \varepsilon \lambda^{\ell}_1(\vec x^1), \quad \ell=1:L
\end{equation}
where
\begin{equation}
    \left(\lambda^\ell_0(\vec x^0)\right)^T = \left. \frac{\partial f}{\partial f^\ell}\right|_{\vec x^0} = \left(\lambda^\ell(\vec x^0)\right)^T = W^{L+1} \prod_{k=\ell+1}^L\left( {\rm diag}\left(D^k_0(\vec x^0)\right) W^k\right).
\end{equation}
In particular, we have the following recursive definitions of $\lambda^{\ell}_0(\vec x^0)$ and $\lambda^{\ell}_1(\vec x^1)$:
\begin{equation}
    \left(\lambda^\ell_0(\vec x^0)\right)^T  = \left(\lambda^{\ell+1}_0(\vec x^0)\right)^T {\rm diag}\left(D^{\ell+1}_0(\vec x^0)\right) W^{\ell+1}
\end{equation}
and
\begin{equation}
\begin{aligned}
    \left(\lambda^\ell_1(\vec x^1)\right)^T &=  \left(\lambda^\ell_1(\vec x^1;\theta,\varepsilon)\right)^T \\
    &= \left(\lambda^{\ell+1}_1(\vec x^1)\right)^T{\rm diag}(D_0^{\ell+1}(\vec x^0))W^{\ell+1} \\
    &+  \left(\lambda_0^{\ell+1}(\vec x^0)\right)^T \left( {\rm diag}\left(D^{\ell+1}_1(\vec x^1)\right) W^{\ell+1}\right) \\
    &+ \varepsilon \left(\lambda^{\ell+1}_1(\vec x^1)\right)^T {\rm diag}\left(D^{\ell+1}_1(\vec x^1)\right) W^{\ell+1},
\end{aligned}
\end{equation}
with 
\begin{equation}
    \left(\lambda^L_0(\vec x^0)\right)^T = W^{L+1} \quad \text{and} \quad \lambda^L_1(\vec x^1) = 0.
\end{equation}
Furthermore, we have the explicit upper bound for $\left\| \lambda^\ell_1 \right\|$ as
\begin{equation}
    \left\| \lambda^\ell_1 \right\| \le \left\|W^{L+1}\right\| \prod_{k=\ell+1}^L \left\|W^k\right\|  \left( \sum_{\ell+1 \le i \le L} \|D^i_1\| \prod_{\ell+1 \le k\neq i\le L} \left\| D_0^k\right\|\right) + \mathcal O\left(\varepsilon\right). 
\end{equation}
\end{lemma}

\begin{proof}
We prove it by induction.
For $\lambda^L$, by definition, we have
\begin{equation}
    \left(\lambda^L(\vec x)\right)^T = \left(\frac{\partial f}{\partial f^{L}}\right)^T = W^{L+1}.
\end{equation}
To make~\Cref{eq:ts-lambda} holds for $\ell=L$, we define $\left(\lambda^L_0(\vec x^0)\right)^T = W^{L+1}$ and $\lambda^L_1(\vec x^1) = 0$.
Thus, for any $1 \le \ell <L$, 
    \begin{equation}
        \left(\lambda^\ell(\vec x)\right)^T =
        \left(\frac{\partial f}{\partial f^{\ell+1}} \cdot 
        \frac{\partial f^{\ell+1}}{\partial f^{\ell}}\right)^T 
        = \left(\lambda^{\ell+1}(\vec x)\right)^T{\rm diag}\left(D^{\ell+1}(\vec x)\right) W^{\ell+1}.
    \end{equation}

    Given the decomposition of $D^{\ell+1}(\vec x)$ in Lemma~\ref{lemm:Dl} and $\lambda^{\ell+1}(\vec x)$ in~\Cref{eq:ts-lambda}, we have
    \begin{equation}
        \begin{aligned}
            &\left(\lambda^\ell(\vec x)\right)^T \\
            = &\left(\lambda^{\ell+1}_0(\vec x^0) + \varepsilon \lambda^{\ell+1}_1(\vec x^1)\right)^T {\rm diag}\left(D^{\ell+1}_0(\vec x^0) + \varepsilon D^{\ell+1}_1(\vec x^1)\right) W^{\ell+1} \\
            = &\left(\lambda^{\ell+1}_0(\vec x^0) \right)^T {\rm diag}\left(D^{\ell+1}_0(\vec x^0) \right) W^{\ell+1}  \\
            &+\varepsilon \left(\left(\lambda^{\ell+1}_1(\vec x^1)\right)^T{\rm diag}(D_0^{\ell+1}(\vec x^0)) \right.\\
            &+  \left(\lambda_0^{\ell+1}(\vec x^0)\right)^T \left( {\rm diag}D^{\ell+1}_1(\vec x^1)\right)  \\
            &+\left. \varepsilon \left(\lambda^{\ell+1}_1(\vec x^1)\right)^T {\rm diag}\left(D^{\ell+1}_1(\vec x^1)\right)  \right) W^{\ell+1}\\
            =& \lambda^{\ell}_0(\vec x^0) + \varepsilon \lambda^{\ell}_1(\vec x^1;\theta,\varepsilon).
        \end{aligned}
    \end{equation}
    Thus, we obtain the recursive formula for $\lambda^{\ell}_0(\vec x^0)$ and $\lambda^{\ell}_1(\vec x^1)$. To explicitly bound $\lambda^{\ell}_1(\vec x^1)$, we apply the recursion repeatedly and have
    \begin{equation}
        \begin{aligned}
           \left(\lambda^\ell_1\right)^T =  W^{L+1} \sum_{\ell+1 \le i \le L}\widetilde{\prod_{\ell+1 \le k\neq i\le L}}  \left( {\rm diag}\left(D^i_1\right)W^i \left( {\rm diag}\left(D^k_0\right) W^k\right) \right) + \mathcal O\left(\varepsilon\right).
        \end{aligned}
    \end{equation}
Noticing the non-commutativity of the multiplication of matrices, we denote
\begin{equation}
\begin{aligned}
    &\widetilde{\prod_{\ell+1 \le k\neq i\le L}}  \left( {\rm diag}\left(D^i_1\right)W^i \left( {\rm diag}\left(D^k_0\right) W^k\right) \right)\\
    =
    &{\rm diag}\left(D^L_0\right)W^L\cdots \\ &{\rm diag}\left(D^{i+1}_0\right)W^{i+1}{\rm diag}\left(D_1^i\right)W^i{\rm diag}\left(D_0^{i-1}\right)W^{i-1}\\&\cdots
    {\rm diag}\left(D^{\ell+1}_0\right)W^{\ell+1}.
\end{aligned}
\end{equation}
Thus, we can bound $\|\lambda^\ell_1\|$ by
\begin{equation}
\begin{aligned}
    \|\lambda^\ell_1\| \le\left\|W^{L+1}\right\| \prod_{k=\ell+1}^L \left\|W^k\right\|  \left( \sum_{\ell+1 \le i \le L} \|D^i_1\| \prod_{\ell+1 \le k\neq i\le L} \left\| D_0^k\right\|\right) + \mathcal O\left(\varepsilon\right).
\end{aligned}
\end{equation}

\end{proof}

\begin{corollary}
For $\widetilde \lambda^\ell(\vec x) = \lambda^\ell(\vec x) \odot D^\ell(\vec x) \in \mathbb R^{n_\ell}$, we have
\begin{equation}
    \widetilde \lambda^\ell(\vec x) = \widetilde \lambda^\ell_0(\vec x^0) + \varepsilon \widetilde \lambda^\ell_1(\vec x^1)
\end{equation}
where
\begin{equation}
   \widetilde \lambda^\ell_0(\vec x^0) = \lambda^\ell_0(\vec x^0) \odot D_0^\ell(\vec x^0) = \widetilde \lambda^\ell(\vec x^0)
\end{equation}
and
\begin{equation}
    \left\| \widetilde \lambda^\ell_1(\vec x^1) \right\| \le \|\lambda_0^1(\vec x^0)\|\|D_1^\ell(\vec x^1)\| + \|\lambda_1^\ell(\vec x^1)\|\|D_0^\ell(\vec x^0)\| + \mathcal O(\varepsilon).
\end{equation}
\end{corollary}
\begin{proof}
    Given Lemma~\ref{lemm:ts-lambdal} of $\lambda^\ell$ and Lemma~\ref{lemm:Dl} of $D^\ell$, we have 
    \begin{equation}
        \begin{aligned}
            \widetilde \lambda^\ell(\vec x) &= \lambda^\ell(\vec x) \odot D^\ell(\vec x) \\
            &= \left(\lambda^\ell_0(\vec x^0) + \varepsilon \lambda^\ell_1(\vec x^1)\right) \odot \left(D_0^\ell(\vec x^0)+\varepsilon D_1^\ell(\vec x^1)\right) \\
            &=\lambda^\ell_0(\vec x^0) \odot D_0^\ell(\vec x^0) + \varepsilon\left(\lambda_0^1(\vec x^0)\odot D_1^\ell(\vec x^1) + \lambda_1^\ell(\vec x^1) \odot D_0^\ell(\vec x^0) + \varepsilon \lambda_1^\ell(\vec x^1)\odot D_1^\ell(\vec x^1) \right) \\
            &=\widetilde \lambda^\ell_0(\vec x^0) + \varepsilon \widetilde \lambda^\ell_1(\vec x^1;\theta,\varepsilon) .
        \end{aligned}
    \end{equation}
Consequently, we have 
\begin{equation}
\begin{aligned}
    \left\|\widetilde \lambda^\ell_1(\vec x^1) \right\| &= \left\| \lambda_0^1(\vec x^0)\odot D_1^\ell(\vec x^1) + \lambda_1^\ell(\vec x^1) \odot D_0^\ell(\vec x^0) + \varepsilon \lambda_1^\ell(\vec x^1)\odot D_1^\ell(\vec x^1)\right\| \\
    &\le \|\lambda_0^1(\vec x^0)\|\|D_1^\ell(\vec x^1)\| + \|\lambda_1^\ell(\vec x^1)\|\|D_0^\ell(\vec x^0)\| + \mathcal O(\varepsilon).
\end{aligned}
\end{equation}
\end{proof}

\subsubsection{Two-scale structure in dataset $\mathcal D$}
Before we study the two-scale structure of the gradient flow of DNNs on dataset $\mathcal D$, we also need to show the two-scale structure of the dataset depending on the two-scale structure of $\vec x$ and the properties of the target $g(\vec x)$.
\begin{lemma}\label{lem:decompose-g}
    For any $g \in C^1(\Omega)$ with $\Omega \subset \mathbb R^d$, and $\vec x$ with two scales under Assumption \ref{assumption:1}, there exists a continuous function $g_1: \mathbb R^d \mapsto \mathbb R$ such that
    \begin{equation}
        g(\vec x) = g_0(\vec x^0) + \varepsilon g_1(\vec x^1),
    \end{equation}
    where $g_0(\vec x^0) = g(\vec x^0)$ and $\vec x^1 = \vec x$.
\end{lemma}
\begin{proof}
    Use the intermediate value theorem and similar arguments shown in the proofs of Lemma~\ref{lemm:fl} and Lemma~\ref{lemm:Dl}.
\end{proof}
Given the above decomposition for the target function, we introduce the following notation
\begin{equation}
\begin{aligned}
    \left. \frac{\partial \mathcal L}{\partial W^{\ell}} \right|_{\mathcal D^0} &:= \frac{\partial}{\partial W^\ell}\left( \frac{1}{N}\sum_{(\vec x, g) \in \mathcal D^0} \left(f(\vec x;\theta) - g\right)^2\right) \\
    &= \frac{\partial}{\partial W^\ell}\left( \frac{1}{N}\sum_{i=1}^N \left(f(\vec x^0_i;\theta) - g(\vec x^0_i)\right)^2\right)\\
    &= \left< \left(f(\vec x^0)-g(\vec x^0)\right)\widetilde \lambda^\ell(\vec x^0) \left(f^{\ell-1}(\vec x^0)\right)^T \right>_{\mathcal D^0}.
\end{aligned}
\end{equation}

\subsubsection{Two-scale structure in $\frac{\partial \mathcal L}{\partial W^\ell}$.}
Based on the previous decompositions and estimates, we can finally present the estimate for $\frac{\partial \mathcal L}{\partial W^{\ell}}$ as follows:
\begin{theorem}\label{them:Al}
For any DNN defined in~\Cref{eq:defdnn} and dataset $\mathcal D$ with two scales as in~\Cref{eq:Dtwoscale} under Assumption~\ref{assumption:1}, we have 
\begin{equation}
\frac{\partial \mathcal L}{\partial W^{\ell}} = \left<A^{\ell}_0(\vec x^0)\right>_N + \varepsilon \left<A_1^{\ell}(\vec x^1)\right>_N,
\end{equation}
 where 
\begin{equation}
    \left<A^{\ell}_0(\vec x^0)\right>_N = \left< \left(f(\vec x^0)-g(\vec x^0)\right)\widetilde \lambda^\ell(\vec x^0) \left(f^{\ell-1}(\vec x^0)\right)^T \right>_{\mathcal D^0} = \left. \frac{\partial \mathcal L}{\partial W^{\ell}} \right|_{\mathcal D^0}
\end{equation}
and 
\begin{equation}
\begin{aligned}
    \left\|A^\ell_1(\vec x^1)\right\| \le  
    &\left|f_1(\vec x^1) - g_1(\vec x^1)\right| \left\|\widetilde \lambda^\ell(\vec x^0)\right\|  \left\|f^{\ell-1}(\vec x^0)\right\| \\
    &+ \left|f(\vec x^0)-g(\vec x^0)\right| \left\|\widetilde \lambda^\ell_1(\vec x^1)\right\|  \left\|f^{\ell-1}(\vec x^0)\right\| \\
    &+ \left|f(\vec x^0)-g(\vec x^0)\right| \left\|\widetilde \lambda^\ell(\vec x^0)\right\|  \left\|f^{\ell-1}_1(\vec x^1)\right\|  + \mathcal O(\varepsilon).
\end{aligned}
\end{equation}
More precisely, we define
\begin{equation}
\begin{aligned}
    \left<A_1^{\ell}(\vec x^1)\right>_N := 
    &\left<\left(f_1(\vec x^1) - g_1(\vec x^1)\right) \widetilde \lambda^\ell(\vec x^0) \left(f^{\ell-1}(\vec x^0)\right)^T \right>_N \\
    &+ \left<\left(f(\vec x^0) - g(\vec x^0) \right) \widetilde \lambda^\ell_1(\vec x^1) \left(f^{\ell-1}(\vec x^0)\right)^T \right>_N \\
    &+ \left<\left(f(\vec x^0) - g(\vec x^0)\right) \widetilde \lambda^\ell(\vec x^0) \left(f_1^{\ell-1}(\vec x^1)\right)^T \right>_N  + \left<\mathcal O(\varepsilon)\right>_{\mathcal D^1}
\end{aligned}
\end{equation}
\end{theorem}
\begin{proof}By definition, we have
\begin{equation}
    \begin{aligned}
        \frac{\partial \mathcal L}{\partial W^{\ell}} 
        &= \left< (f(\vec x)-g) \widetilde \lambda^{\ell}(\vec x) \left(f^{\ell-1}(\vec x)\right)^T \right>_N \\
        &= \left< (W^{L+1}f^L_0 + \varepsilon W^{L+1}f^L_1-g_0 - \varepsilon g_1)\left(\widetilde \lambda^\ell_0 + \varepsilon \widetilde \lambda^\ell_1\right) \left(f^{\ell-1}_0 + \varepsilon f^{\ell-1}_1\right)^T\right>_N   \\
        &= \left< A^{\ell}_0(\vec x^0) \right>_N + \varepsilon \left< A^{\ell}_1(\vec x^1) \right>_N,
    \end{aligned}
\end{equation}
where 
\begin{equation}
\begin{aligned}
    A^{\ell}_0(\vec x^0) &=  (W^{L+1}f^L_0-g_0)\widetilde \lambda^\ell_0 \left(f^{\ell-1}_0\right)^T  \\
    &= \left(f(\vec x^0)-g(\vec x^0)\right)\widetilde \lambda^\ell(\vec x^0) \left(f^{\ell-1}(\vec x^0)\right)^T 
\end{aligned}
\end{equation}
and 
\begin{equation}
    \begin{aligned}
    A_1^\ell(\vec x^1) = & \left(W^{L+1}f^L_1(\vec x^1) - g_1(\vec x^1)\right) \widetilde \lambda^\ell_0(\vec x^0)  \left(f^{\ell-1}_0(\vec x^0)\right)^T \\
    &+ \left(W^{L+1}f^L_0(\vec x^0)-g_0(\vec x^0)\right) \widetilde \lambda^\ell_1(\vec x^1)  \left(f^{\ell-1}_0(\vec x^0)\right)^T \\
    &+ \left(W^{L+1}f^L_0(\vec x^0)-g_0(\vec x^0)\right) \widetilde \lambda^\ell_0(\vec x^0)  \left(f^{\ell-1}_1(\vec x^1)\right)^T  +  \mathcal O(\varepsilon) \\
    = &\left(f_1(\vec x^1) - g_1(\vec x^1)\right) \widetilde \lambda^\ell(\vec x^0)  \left(f^{\ell-1}(\vec x^0)\right)^T \\
    &+ \left(f(\vec x^0)-g(\vec x^0)\right) \widetilde \lambda^\ell_1(\vec x^1)  \left(f^{\ell-1}(\vec x^0)\right)^T \\
    &+ \left(f(\vec x^0)-g(\vec x^0)\right) \widetilde \lambda^\ell(\vec x^0)  \left(f^{\ell-1}_1(\vec x^1)\right)^T  +  \mathcal O(\varepsilon).
    \end{aligned}
\end{equation}
As a result, we have the definition of $\left<A_1^{\ell}(\vec x^1)\right>_N$ and the bound for $\left\|A_1^{\ell}(\vec x^1)\right\|$.

\end{proof}

\begin{remark}
    With a similar fashion, we also have the two-scale structure in $\frac{\partial \mathcal L}{\partial b^\ell}$ since
    \begin{equation}
        \frac{\partial \mathcal L}{\partial b^\ell} = \left< (f(\vec x;\theta)-g)  \widetilde \lambda^\ell(\vec x) \right>_N.
    \end{equation}
\end{remark}

\begin{remark}
Here, we notice that $\mathcal L$ can be thought as a function of $\varepsilon$ since $\vec x_i = \begin{pmatrix}
    x^0_i \\ \varepsilon x^1_i
\end{pmatrix}$.
Thus, let us define 
\begin{equation}
F^\ell(\varepsilon) = \frac{\partial \mathcal L(\theta;\varepsilon)}{\partial W^{\ell}} \in \mathbb R^{n_\ell \times n_{\ell-1}}.
\end{equation}
Then, one may apply the Taylor expansion in terms of $\varepsilon$ to $F^\ell(\varepsilon)$ and get
\begin{equation}
    F^\ell(\varepsilon) = F^\ell(0) + \varepsilon \frac{d F^\ell(0)}{d \varepsilon} + \mathcal O(\varepsilon^2).
\end{equation}
Here, we have
\begin{equation}
    \left<A^\ell_0(x)\right>_N = F^\ell(0), \quad \lim_{\varepsilon \to 0} \left<A_1^\ell\right>_N = \frac{d F^\ell(0)}{d \varepsilon}.
\end{equation}
Our previous analysis gives more precise structures and estimates for $F^\ell(0)$ and $\frac{d F^\ell(0)}{d \varepsilon}$. More important, 
 the direct Taylor expansion for $F^\ell(\varepsilon)$ cannot be generalized to the multiscale case. 
\end{remark}

\subsection{Activation functions that fit the previous analysis.}
According to the previous analysis, we see that our previous analysis can be applied directly if ${\rm Lip}(\sigma)$ and ${\rm Lip}(\sigma')$ are uniformly bounded.
For example:
\begin{description}
    \item[Sigmoid]
    \begin{equation}
        \sigma(t) = \frac{1}{1+e^{-t}}
    \end{equation}
    \item[Tanh]
    \begin{equation}
        \sigma(t) = \frac{e^t - e^{-t}}{e^{t}+e^{-t}} 
    \end{equation}
    \item[Gaussian Error Linear Unit (GeLU)]
    \begin{equation}
        \sigma(t) = t \Phi(t) = t\int_\infty^t \frac{1}{\sqrt{2\pi}} e^{-\frac{s^2}{2}}ds
    \end{equation}
    \item[Softplus]
    \begin{equation}
        \sigma(t) = \ln(1+e^{t}) 
    \end{equation}
    \item[Sigmoid linear unit (SiLU, Sigmoid shrinkage, SiL, or Swish-1)]
    \begin{equation}
        \sigma(t) = \frac{t}{1+e^{-t}}
    \end{equation}
\end{description}
In addition, our previous analysis can also be applied to activation functions belonging to $W^{2,\infty}_{loc}$ since we can assume $\vec x \in \Omega$, which is a bounded set and also $\|\theta\|_{\ell^2} \le M$ in the training process.

\subsection{Multiscale structure in DNNs}
The key to proving the result is the extension of the multiscale decomposition for $f^\ell(\vec x)$ and $D^\ell(\vec x)$ (the result for $\widetilde \lambda^\ell$ can be derived naturally based on these two results). Similar as before, we only outline the least scale components $\vec x^k$ for clarity.
\begin{lemma}\label{lem:mscale-DNN}
For any DNN defined in~\Cref{eq:defdnn} and dataset under Assumption~\ref{assumption:1}, we have
\begin{equation}
    f^\ell(\vec x^k) = f^\ell(\vec x^{k-1}) + \varepsilon^k f^\ell_k(\vec x^k),
\end{equation}
and
\begin{equation}
    D^\ell(\vec x^k) = D^\ell(\vec x^{k-1}) + \varepsilon^k D^\ell_k(\vec x^k),
\end{equation}
for any $\ell=1:L$ and $k=1:m$.
Moreover, we have
\begin{equation}
    \|f^{\ell}_k(\vec x^k)\| \le \left({\rm Lip}(\sigma)\right)^{\ell} \left\|W^1_k x^k \right\| \prod_{j=2}^\ell \left\|W^j\right\| ,
\end{equation} and
\begin{equation}
    \|D^\ell_k(\vec x^k)\| \le {\rm Lip}(\sigma') \left({\rm Lip}(\sigma)\right)^{\ell-1} \left\|W^1_k x^k \right\| \prod_{j=2}^\ell \left\|W^j\right\|.
\end{equation}
where ${\rm Lip}(\sigma)$ and ${\rm Lip}(\sigma')$ denote the Lipschitz constants of activation function $\sigma$ and its derivative $\sigma'$.
\end{lemma}
\begin{proof}
    This is an extension of the Lemma~\ref{lemm:fl}.
\end{proof}

\begin{lemma}
For any DNN defined in~\Cref{eq:defdnn} and dataset under Assumption~\ref{assumption:1}, we have
\begin{equation}
    \widetilde{\lambda}^\ell(\vec x^k) = \widetilde\lambda^\ell(\vec x^{k-1}) + \varepsilon^k \widetilde\lambda^\ell_k(\vec x^k),
\end{equation}
where $\widetilde\lambda^\ell_k(\vec x^k) = \widetilde\lambda^\ell_k(\vec x^k;\theta, \varepsilon^k)$. 
\end{lemma}

Similar to the two-scale decomposition of the target function in Lemma~\ref{lem:decompose-g}, we have the following result for $g(\vec x^k)$.
\begin{lemma}
    For any $g \in C^1(\Omega)$ with $\Omega \subset \mathbb R^d$, and $\vec x^m$ with multiscale structure under Assumption \ref{assumption:1}, there exists a continuous function $g_k: \mathbb R^d \mapsto \mathbb R$ such that
    \begin{equation}
        g(\vec x^k) = g(\vec x^k) + \varepsilon^k g_k(\vec x^k) \quad k=1:m.
    \end{equation}
\end{lemma}
Then, we define the following sets
\begin{equation}
    \mathcal D^k = \left\{ (\vec x_i^{k-1}, g(\vec x^{k-1}_i)) \right\} \quad \text{and}, \quad k=0:m.
\end{equation}
Given these new sets, we introduce the following notation
\begin{equation}
\begin{aligned}
    \left. \frac{\partial \mathcal L}{\partial W^{\ell}} \right|_{\mathcal D^k} &:= \frac{\partial}{\partial W^\ell}\left( \frac{1}{N}\sum_{(\vec x, g) \in \mathcal D^k} \left(f(\vec x;\theta) - g\right)^2\right) \\
    &= \frac{\partial}{\partial W^\ell}\left( \frac{1}{N}\sum_{i=1}^N \left(f(\vec x^k_i;\theta) - g(\vec x^k_i)\right)^2\right)\\
    &= \left< \left(f(\vec x^k)-g(\vec x^k)\right)\widetilde \lambda^\ell(\vec x^k) \left(f^{\ell-1}(\vec x^k)\right)^T \right>_{N},
\end{aligned}
\end{equation}
for all $k=0:m$. Here, we notice that $\frac{\partial \mathcal L}{\partial W^{\ell}} = \left. \frac{\partial \mathcal L}{\partial W^{\ell}} \right|_{\mathcal D} = \left. \frac{\partial \mathcal L}{\partial W^{\ell}} \right|_{\mathcal D^m}$.

Before the show our last theorem about the multiscale structure of the gradient flow of DNNs, we first present the following recursive decomposition for $\frac{\partial \mathcal L(\theta)}{\partial W^\ell}$.
\begin{lemma}\label{lem:mstructure-re}
    For any DNN defined in~\Cref{eq:defdnn} and dataset under Assumption~\ref{assumption:1}, we have
\begin{equation}
    \left. \frac{\partial \mathcal L(\theta)}{\partial W^\ell} \right|_{\mathcal D^k} =  \left. \frac{\partial \mathcal L(\theta)}{\partial W^\ell} \right|_{\mathcal D^{k-1}} + \varepsilon^k\left<  A^\ell_k\left( \vec x^k \right)\right>_N \quad k=1:m,
\end{equation}
where
\begin{equation}\label{eq:Akxk}
\begin{aligned}
    \left<A_k^{\ell}(\vec x^k)\right>_N := 
    &\left<\left(f_k(\vec x^k) - g_k(\vec x^k)\right)\widetilde \lambda^\ell(\vec x^{k-1}) \left(f^{\ell-1}(\vec x^{k-1})\right)^T \right>_N \\
    &+ \left<\left(f(\vec x^{k-1}) - g(\vec x^{k-1})\right) \widetilde \lambda^\ell_1(\vec x^k) \left(f^{\ell-1}(\vec x^{k-1})\right)^T \right>_N \\
    &+ \left<\left(f(\vec x^{k-1}) - g(\vec x^{k-1})\right) \widetilde \lambda^\ell(\vec x^{k-1}) \left(f_k^{\ell-1}(\vec x^k)\right)^T \right>_N \\
    &+ \left<\mathcal O(\varepsilon^k)\right>_N.
\end{aligned}
\end{equation}
More precisely, we have
\begin{equation}\label{eq:Oepsilonk}
    \begin{aligned}
        \mathcal O(\varepsilon^k) = \varepsilon^k&\left( \left(f(\vec x^{k-1}) - g(\vec x^{k-1})\right)\widetilde \lambda^\ell_k(\vec x^{k}) \left(f^{\ell-1}_k(\vec x^{k})\right)^T \right. \\
        &+\left(f_k(\vec x^k) - g_k(\vec x^k)\right)\widetilde \lambda^\ell(\vec x^{k-1}) \left(f^{\ell-1}_k(\vec x^{k})\right)^T \\
        &+ \left. \left(f_k(\vec x^k) - g_k(\vec x^k)\right)\widetilde \lambda^\ell_k(\vec x^{k}) \left(f^{\ell-1}(\vec x^{k-1})\right)^T \right) \\
        +\varepsilon^{2k} &\left(f_k(\vec x^k) - g_k(\vec x^k)\right)\widetilde \lambda^\ell_k(\vec x^{k}) \left(f^{\ell-1}_k(\vec x^{k})\right)^T,
    \end{aligned}
\end{equation}
for all $k=1:m$.
\end{lemma}

\begin{proof}
By definition and the decomposition for $f(\vec x^k), g(\vec x^k), \widetilde \lambda^\ell(\vec x^k)$, and $f^{\ell-1}(\vec x^k)$, we have
\begin{equation}
\begin{aligned}
    \left. \frac{\partial \mathcal L}{\partial W^{\ell}} \right|_{\mathcal D^k} = &\left< \left(f(\vec x^k)-g(\vec x^k)\right)\widetilde \lambda^\ell(\vec x^k) \left(f^{\ell-1}(\vec x^k)\right)^T \right>_{N} \\
    = &\left< \left(f(\vec x^{k-1})-g(\vec x^{k-1}) + \varepsilon^k(f_k(\vec x^{k})-g_k(\vec x^{k-1}))\right) \right. \\
    &\left. \left( \widetilde\lambda^\ell(\vec x^{k-1}) + \varepsilon^k \widetilde\lambda^\ell_k(\vec x^k)\right)\left(f^{\ell-1}(\vec x^{k-1}) + \varepsilon^kf^{\ell-1}_k(\vec x^{k})\right)^T \right>_{N} \\
    = &\left<\left(f(\vec x^{k-1})-g(\vec x^{k-1})\right)\widetilde \lambda^\ell(\vec x^{k-1}) \left(f^{\ell-1}(\vec x^{k-1})\right)^T\right>_N +  \varepsilon^k\left<  A^\ell_k\left( \vec x^k \right)\right>_N \\
    = &\left. \frac{\partial \mathcal L(\theta)}{\partial W^\ell} \right|_{\mathcal D^{k-1}} + \varepsilon^k\left<  A^\ell_k\left( \vec x^k \right)\right>_N.
\end{aligned}
\end{equation}
    
\end{proof}

Finally, we have the following main theorem about the multiscale structure of the gradient flow of DNNs.

\begin{theorem}
For any DNN defined in~\Cref{eq:defdnn} and dataset under Assumption~\ref{assumption:1}, we have
\begin{equation}
    \frac{\partial \mathcal L(\theta)}{\partial W^\ell} = \sum_{k=0}^m \varepsilon^k\left<  A^\ell_k\left( \vec x^k \right)\right>_N,
\end{equation}
where
\begin{equation}
    A^{\ell}_0(\vec x^0) = \left(f(\vec x^0)-g(\vec x^0)\right)\widetilde \lambda^\ell(\vec x^0) \left(f^{\ell-1}(\vec x^0)\right)^T
\end{equation}
and $\left<  A^\ell_k\left( \vec x^k \right)\right>_N$ follow the definitions in~\Cref{eq:Akxk} and~\Cref{eq:Oepsilonk} in Lemma~\ref{lem:mstructure-re} for $k=1:m$.
\end{theorem}
\begin{proof}
    This is a direct result of applying Lemma~\ref{lem:mstructure-re} repeatedly.
\end{proof}
We also want to point out that same structure of multiscale expansion occurs in other models: for linear regression, we have 
\begin{equation}
    \frac{d \theta}{d t} = \sum_{i=0}^d \varepsilon^i \left< A_i\left( \vec  x^i \right)\right>_N,
\end{equation}
where
\begin{equation}
    \left< A_i\left( \vec  x^i \right)\right>_N = \left<\vec  x^i \left(\vec  x^i\right)^T\right>_N - \left<\vec  x^{i-1} \left(\vec  x^{i-1}\right)^T\right>_N,
\end{equation}
with $\left< A_0\left( \vec  x^0 \right)\right>_N = \left<\vec  x^0 \left(\vec  x^0\right)^T\right>_N$.

For logistic regression, we still have
\begin{equation}
    \frac{d \theta}{d t} = \sum_{i=0}^d \varepsilon^i \left< A_i\left( \vec  x^i \right) \right>_N.
\end{equation}
\section{Proof of multiscale gradient components and multiscale Hessian:}
\subsection{Multiscale gradient components in logistic regression}\label{proof:multiscale_grad_logit}
\begin{proof}
The label $g(\vec x_i)\in \bbR^k$ is a probability. Let $\vec p_i := g(\vec x_i)$, we have $\sum_{j=1}^k p_{ij}=1$, and $p_{ij}>0$.  Then, the cross-entropy loss can be explicitly written as:
\[ \cL_c = -\frac{1}{N}\sum^N_{i=1} \vec p_i \cdot \log\big( \vec f(\vec x_i; \vec w,b) \big)=  -\frac{1}{N}\sum^N_{i=1} \sum_{j=1}^k p_{ij} \log\big( [f(\vec x_i; \vec w,b)]_{j} \big).  \]
Recall that:
\begin{equation*}
        \left[f({\vec  x; \vec w, b})\right]_j  = \frac{e^{\vec w_j \cdot \vec  x + b_j}}{\sum_{i=1}^k e^{\vec w_i \cdot \vec  x + b_i}}, \quad j = 1:k,
    \end{equation*}
and note $\log(f_j) = (\vec w_j\cdot \vec x+b_j) - \log(\sum_i e_i)$, with $e_j:=e^{\vec w_j\cdot\vec x+b_j}, \, f_j:= [f(\vec x_i; \vec w,b)]_{j}$, for clarity. Therefore, for any weight $\vec w_l$: 
\[ \pp{\log(f_j)}{\vec w_l} = (\vec \delta_{jl} - \frac{e_l}{\sum_i e_i})\vec x^T\]
Then:
\begin{equation}\label{eqn:grad_logit_proof}
     \pp{\cL_c}{\vec w_l} = -\frac{1}{N}\sum_{i=1}^N \sum_{j=1}^k  p_{ij}  (\delta_{jl} - \frac{e_l}{\sum_i e_i})\vec x_i^T =  -\frac{1}{N}\sum_{i=1}^N \vec x_i^T  (p_{il}  - \frac{e_l}{\sum_i e_i}) = \left < \big(g(\vec x)_l - f(\vec x)_l\big) \vec x^T \right >_N .
\end{equation}


where the multiplier $\left( g(\vec x_i)_l -f({\vec  x_i; \vec w, b})_l \right) \sim\cO(1)$ in general. Therefore, it is straightforward to see that components in the loss gradient
exhibits a similar multiscale behavior as the dataset, where the magnitude of the gradient aligns with the scale of the data distribution:
\begin{equation}\label{eqn:multiscale_grad_proof}
    \frac{\partial \mathcal L_c}{\partial \vec w_l} \sim \big.\left(\cO(1), \cO(\varepsilon_1), \cO(\varepsilon_2)\dots \big.\right).
\end{equation}
Organizing the gradient for each $\vec w_l$ into the matrix form leads us to the desired result.
\end{proof}

\subsection{Multiscale gradient components in neural network}\label{proof:multiscale_grad_NN}
\begin{proof}
    The gradient of the loss concerning the first layer parameter $W^1$ is given by:
\[ \pp{\cL}{W^1} = \frac{1}{2N}\sum_{i=1}^N \left( f(\vec x_i, W) -g_i \right) \pp{f}{W^1},\]
where from the multiplicative representation:
\[\pp{f}{W^1}= \pp{\tilde f}{W^1} = \pp{\tilde f(W^1\vec x, W_2)}{W^1\vec x} \vec x^T. \]
Without loss of generality, we focus on the gradient of the first neuron, and let $\vec w_{11}$ denote the first row of the weight matrix $W^1$, such that $\vec w_{11}\cdot \vec x$ produces the first entry in the first hidden layer. Similarly, we have 
\[\pp{f}{\vec w_{11}} = \pp{\tilde f}{\vec w_{11}}  = \pp{\tilde f(W^1\vec x, W_2)}{\vec w_{11}^T\vec x} \vec x^T. \]
Since $f$ is a scalar function and $\vec w_{11}^T\vec x$ is also a scalar, we set $c_{1}:=  \pp{\tilde f(W^1\vec x, W_2)}{\vec w_{11}^T\vec x} \in \bbR$ for simplicity, which yields $\pp{f}{\vec w_{11}} = c_{1}\vec x^T$, and the same applies for any row of the weight matrix $W^1$. Let $c_i$ denote the corresponding coefficient for the $i$-th row, and we have $\pp{\tilde{f}}{W^1} = \vec c \vec x^T$
where $\vec c = [c_1, c_2, \dots ]^T$. Plug in the expression back to the loss gradient produces:
\begin{equation}\label{eqn:grad_NN_proof}
    \pp{\cL}{\vec w_{11}} = \frac{1}{2N}\sum_{i=1}^N \left( f(\vec x_i, W) -g_i \right) c_1(\vec x_i) \vec x_i^T.
\end{equation}
The gradient expression in \Cref{eqn:grad_NN_proof} holds for any neuron in the first layer and closely matches with~\Cref{eqn:grad_logit_proof}. Thus, we can deduce that the gradient for $W^1$ also explicitly relies on the data distribution, where, if the data distribution satisfies~\Cref{assumption:1}, the gradient will have dimension-wise multiscale characteristics, akin to \Cref{eqn:multiscale_grad_proof}.

\end{proof}

\subsection{Multiscale Hessian in neural network}\label{proof:multiscale_Hess_NN}
\begin{proof}
    Continued from~\Cref{eqn:grad_NN_proof}, to derive the Hessian we have:
\begin{align*}
   \nabla^2_{\vec w_{11}} \cL:  =&  \frac{1}{2N}\sum_{i=1}^N  \pp{f(\vec x_i)}{\vec w_{11}} c_1(\vec x_i) \vec x_i^T + \left( f(\vec x_i, W) -g_i \right) \nabla^2_{\vec w_{11}} f(\vec x_i) \\
    =& \frac{1}{2N}\sum_{i=1}^N c_1(\vec x_i)^2 \vec x_i\vec x_i^T +\left( f(\vec x_i, W) -g_i \right) \pp{^2 \tilde{f}(W^1\vec x_i, W_2)}{(\vec w_{11}^T\vec x)^2}\vec x_i\vec x_i^T.
\end{align*}

Note that $\left( f(\vec x_i, W) -g_i \right), \, \pp{^2 \tilde{f}(W^1\vec x, W_2)}{(\vec w_{11}^T\vec x_i)^2}\in\bbR$, setting: 
\[s_j(\vec x_i) = c_j(\vec x_i)^2+ \left( f(\vec x_i, W) -g_i \right)\pp{^2 \tilde{f}(W^1\vec x, W_2)}{(\vec w_{1j}^T\vec x_i)^2}\in\bbR,\]
leads to \[\nabla^2_{\vec w_{1j}} \cL  = \frac{1}{2N}\sum_{i=1}^{N} s_j(\vec x_i) \vec x_i\vec x_i^T \approx \left(\sum_{i=1}^N s_j(\vec x_i)\right) X^TX, \]
which finishes the proof.
\end{proof}


\section{Proof of convergence properties of MrGD:}

\subsection{Proof of Theorem~\ref{thm:Sm} (convergence for quadratic problems): }\label{sec:proofSm}
\begin{lemma}\label{lem:Rkappa}
    For any $i=2:m$ and $j=1:m$, we have the following inequalities
    \begin{equation}
        1 - R_i^j\kappa_{i}^{-1} \ge 1-R_{i-1}^j\kappa_{i-1}^{-1}
    \end{equation}

\end{lemma}
\begin{proof} 
Direct calculation gives:
 \[ 1 - R_i^jk_i^{-1} = 1-\eta_j\sigma_i\frac{\sigma_{i,d_i}}{\sigma_i} = 1-\eta_j\sigma_{i,d_i} \ge 1 - \eta_j\sigma_{i-1,d_{i-1}} =  1-R_{i-1}^j\kappa_{i-1}^{-1}\]
\end{proof}

\paragraph{The Main Proof:}
\begin{proof}
    Let $V_i$ be the eigenspace corresponding to the $i$-th group eigenvalues of $A$, and we can write:
    \begin{equation}
        \mathbb R^d = V = V_1 \oplus V_2 \oplus \cdots \oplus V_m,
    \end{equation}
    since $A$ is symmetric and different groups of eigenvalues are distinct. This decomposition leads to
    \begin{equation}
        \|S\| = \sup_{\|v\|=1} \|Sv\| \le \max_{i=1:m} \sup_{v_i \in V_i, \|v_i\|=1} \|Sv_i\|
    \end{equation}
    since $S$ is a polynomial of $A$ and $\{V_i\}_{i=1}^m$ are also the eigenspaces of $S$, orthogonal to each other.

    Then, the proof idea can be divided into two steps. First, we establish that for any $i=1:m$, there exists some constant $c_i$'s such that
    \begin{equation}\label{eq:findci}
     \sup_{v_{i} \in V_{i},~\|v_{i}\|=1} \|Sv_{i}\| \le c_i.
    \end{equation}
 Subsequently, we demonstrate that by posing suitable conditions on $n_i$'s, these $c_i$'s satisfy:
    \begin{equation}\label{eq:ciorder}
    c_{i+1} \ge c_i, \quad \forall i=1:m-1.
    \end{equation}
    Consequently, if we finish the aforementioned two steps, we can conclude that
    $$
    \|S\| \le c_m,
    $$
\textbf{Step $1$:} Following the discussion from~\Cref{rmk:convergence}, $c_i$'s in~\Cref{eq:findci} can be chosen as:
    \begin{equation}\label{eq:defci}
        \sup_{v_{i} \in V_{i},~\|v_{i}\|=1}\|S v_{i}\| \le  c_i  := \prod_{j=1}^{i} \left(1 - R_i^j\kappa_{i}^{-1}\right)^{n_j}   \prod_{j=i+1}^m \left|1- R^j_i \right|^{n_j}.
    \end{equation}
    In particular, we have
    \begin{equation}
         \sup_{v_{m} \in V_{m},~\|v_{m}\|=1}\|S v_{m}\| = \prod_{j=1}^{m} \left(1 - R_m^j\kappa_{m}^{-1}\right)^{n_j} {=:c_m}.
    \end{equation}

\textbf{Step $2$:} We derive sufficient conditions for~\Cref{eq:ciorder} by induction from $m-1$ to $1$. This process will generate conditions on $n_i$'s along the way.

For $i=m-1$, we want to ensure $c_m\ge c_{m-1}$, that is:
\[  \prod_{j=1}^{m} \left(1 - R_m^j\kappa_{m}^{-1}\right)^{n_j} \ge\prod_{j=1}^{m-1} \left(1 - R_{m-1}^j\kappa_{m-1}^{-1}\right)^{n_j}    \left|1- R^m_{m-1} \right|^{n_m}.    \]
We divide the above into two parts, where the first part is to consider only up to $j=m-2$, and is directly given by Lemma~\ref{lem:Rkappa}:
\begin{equation}
    \prod_{j=1}^{m-2} \left(1 - R_m^j \kappa_m^{-1}\right)^{n_j} \ge 
    \prod_{j=1}^{m-2} \left(1 - R_{m-1}^j\kappa_{m-1}^{-1}\right)^{n_j},
\end{equation}

for arbitrary $n_j > 0$. Thereby, we only need to ensure that the second part holds:
\[ \left(1-R_m^{m-1} \kappa_m^{-1}\right)^{n_{m-1}} \left(1 - R_m^m \kappa_m^{-1}\right)^{n_m}  \ge \left(1-R_{m-1}^{m-1} \kappa_{m-1}^{-1}\right)^{n_{m-1}}\left|1 - R_{m-1}^{m} \right|^{n_m}. \]
A direct calculation yields:
\[     
n_{m-1}\log\left(\frac{1-R_m^{m-1} \kappa_m^{-1}}{1-R_{m-1}^{m-1} \kappa_{m-1}^{-1}}\right) \ge n_m \log\left( \frac{\left|1 - R_{m-1}^{m} \right|}{1 - R_m^m \kappa_m^{-1}}\right).
\]
By setting:
\[  F_{m-1, m} = \left.  \left(-\log(r_{m-1}) +\log\left(\frac{\left|r_{m-1} - R^m_m\right|}{1-R^m_m\kappa_m^{-1}}\right) \right)    \right\slash \log\left(\frac{1-R^{m-1}_m\kappa_m^{-1}}{1-R^{m-1}_{m-1}\kappa_{m-1}^{-1}}\right),\]



we obtain the following sufficient condition for $c_m\ge c_{m-1}$:
\begin{equation}\label{eq:m_d-1*}
    n_{m-1} \ge n_m F_{m-1,m}.
\end{equation}

Inductively, assume $n_{m-1}, n_{m-2}, \cdots, n_{i+1}$ have been properly defined such that
 $ c_{j+1} \ge c_{j} $ holds for all $j=i+1:m-1$, we aim to derive a condition on $n_i$ as the sufficient condition for
\begin{equation}
\label{eq:SvSV}
        c_{i+1} \ge c_i.
    \end{equation}
We again divide~\Cref{eq:SvSV} into two parts, and invoke Lemma~\ref{lem:Rkappa} for the first part up to $j=i-1$:
\begin{equation}
    \prod_{j=1}^{i-1} \left(1 - R_{i+1}^{j}\kappa_{i+1}^{-1}\right)^{n_j} \ge 
    \prod_{j=1}^{i-1} \left(1 - R_{i}^{j}\kappa_{i}^{-1}\right)^{n_j} ,
\end{equation}
for arbitrary $n_j > 0$. The second part needs to satisfy:
\begin{equation}\label{eqn:ci_2nd}
\begin{aligned}
 &\left(1 - R_{i+1}^i\kappa_{i+1}^{-1}\right)^{n_i} \left(1 - R_{i+1}^{i+1}\kappa_{i+1}^{-1}\right)^{n_{i+1}}   \prod_{j=i+2}^m \left|1- R^j_{i+1} \right|^{n_j}\\
 \ge &\left(1 - R_i^i\kappa_{i}^{-1}\right)^{n_i} \left|1 - R_i^{i+1}\right|^{n_{i+1}}  \prod_{j=i+2}^m \left|1- R^j_i \right|^{n_j}.   
\end{aligned}
\end{equation}
We claim the following condition on $n_i$ is sufficient for the purpose,
\begin{equation}\label{eq:sufficient}
    n_i \ge \left\lceil\sum_{j=i+1}^m n_j F_{i,j}\right\rceil,
\end{equation}
where $F_{i,j}$ will be determined shortly. Thus, we only need to check~\Cref{eqn:ci_2nd} by setting $n_i =\sum_{j=i+1}^m n_j F_{i,j}$ whereby the equation can be further divided into two separate parts:
\begin{equation}
   \left(1 - R_{i+1}^i\kappa_{i+1}^{-1}\right)^{n_{i+1}F_{i,i+1}} \left(1 - R_{i+1}^{i+1}\kappa_{i+1}^{-1}\right)^{n_{i+1}} 
 \ge \left(1 - R_i^i\kappa_{i}^{-1}\right)^{n_{i+1}F_{i,i+1}} \left|1 - R_i^{i+1}\right|^{n_{i+1}},  
\end{equation}
and 
\begin{equation}
   \left(1 - R_{i+1}^i\kappa_{i+1}^{-1}\right)^{n_{j}F_{i,j}} \left|1- R^j_{i+1} \right|^{n_j}
 \ge \left(1 - R_i^i\kappa_{i}^{-1}\right)^{n_{j}F_{i,j}}\left|1- R^j_i \right|^{n_j},
\end{equation}
for all $j=i+2:m$. Similarly, a direct calculation gives:
\[  F_{i, i+1} = \left.  \left(-\log(r_{i}) +\log\left(\frac{\left|r_{i} - R^{i+1}_{i+1}\right|}{1-R^{i+1}_{i+1}\kappa_{i+1}^{-1}}\right) \right)    \right\slash \log\left(\frac{1-R^{i}_{i+1}\kappa_{i+1}^{-1}}{1-R^{i}_{i}\kappa_{i}^{-1}}\right),\]
and
\[   F_{i,j} =  \left.\left(-\log(r_{i}) + \log\left( \frac{\left|r_i - R_{i+1}^{j}\right|}{\left|1-R_{i+1}^{j}\right|}\right) \right)
        \right \slash \log\left(\frac{1-R^{i}_{i+1}\kappa_{i+1}^{-1}}{1-R^{i}_{i}\kappa_{i}^{-1}}\right), \]


for all $j=i+2:m$.
\end{proof}


\subsection{Proof for~\Cref{cor:FijGr}}\label{proof:FijGr}
\begin{proof}
First, we notice that the denominators of $F_{i,i+1}$ and $F_{i,j}$ for $j=i+2:m$ given by~\Cref{thm:Sm} are the same and larger than $0$. Therefore, we first give a lower bound on the denominator:
\begin{align*}
    \log\left(\frac{1-R^{i}_{i+1}\kappa_{i+1}^{-1}}{1-R^{i}_{i}\kappa_{i}^{-1}}\right) &=   \log\left(\frac{1-\eta_i \sigma_{i+1,d_{i+1}}}{1-\eta_i\sigma_{i,d_i}}\right) = \log\left(\frac{1-\frac{r}{\eta\kappa_c}}{1-\frac{1}{\eta\kappa_c}}\right)\\ &= \log\left(1-\frac{r}{\eta\kappa_c}\right) - \log\left(1-\frac{1}{\eta\kappa_c}\right).
\end{align*}
Recall the log-inequality:
\[ 1-\frac{1}{x} \le \log(x) \le x-1 \; \text{ for } x>0,\quad \implies \quad \frac{-x}{1-x}\le\log(1-x)\le -x \; \text{ for } x<1.\]
Since $\eta\kappa_c > 1$ and $r < 1$, the log-inequality yields the following bound on the denominator:
\[ \log\left(\frac{1-R^{i}_{i+1}\kappa_{i+1}^{-1}}{1-R^{i}_{i}\kappa_{i}^{-1}}\right) \ge \frac{-\frac{r}{\eta\kappa_c}}{1-\frac{r}{\eta\kappa_c}} + \frac{1}{\eta\kappa_c} = \frac{\eta\kappa_c(1-r)-r}{\eta\kappa_c(\eta\kappa_c-r)}.\]


As for the numerator for $F_{i,i+1}$, since $r_i = r \ll 1$, it satisfies: 
\[ \left(-\log(r_{i}) +\log\left(\frac{\left|r_{i} - R^{i+1}_{i+1}\right|}{1-R^{i+1}_{i+1}\kappa_{i+1}^{-1}}\right) \right) \ge \log\left(\frac{\left|1-\eta_{i+1}\sigma_i\right|}{1-\eta_{i+1}\sigma_{i+1,d_{i+1}}}\right) > 0. \]



Hence, we derive a corresponding upper bound:
\begin{align*}
    \log\left(\frac{\left|1-\eta_{i+1}\sigma_i\right|}{1-\eta_{i+1}\sigma_{i+1,d_{i+1}}}\right) &= \log\left(\frac{ \kappa_c - \eta\kappa_c r}{\eta\kappa_c r - r}\right) = \log(\kappa_c) +\log(1-\eta  r)-\log(r)  - \log(\eta\kappa_c-1) \\
    &\le -\log(r) - \log(\eta\kappa_c - 1) + \kappa_c -\eta r -1.
\end{align*}

Next, for $F_{i,j}$ with $j=i+2:m$, we denote $k=j-(i+1) \ge 1$ and its numerator becomes:
\[ \left(-\log(r_{i}) + \log\left(\frac{\left|r_i - R_{i+1}^{j}\right|}{\left|1-R_{i+1}^{j}\right|}\right) \right) = -\log(r) + \log\left(\frac{1-\eta r^{k+1}}{1 - \eta r^{k}}\right)\ge 0, \]
since $r \ll 1$. Again, we derive a corresponding upper bound by applying the log-inequality:
\[ \log\left(\frac{1-\eta r^{k+1}}{1 - \eta r^{k}}\right) = \log(1-\eta r^{k+1}) - \log(1-\eta r^{k}) \le -\eta r^{k+1} + \frac{\eta r^k}{1-\eta r^k} \sim o(r^k). \]
Therefore, there is a constant $C>0$ (depends on $k$, $\varepsilon$, and $\eta$) \emph{s.t.} the numerator is bounded above:
\begin{equation}
   \log\left(\frac{1-\eta r^{k+1}}{1 - \eta r^{k}}\right) \le Cr^k,
\end{equation}
for any $r \in (0, \varepsilon)$. This finishes the proof.





\end{proof}

\subsection{Proof of Theorem~\ref{thm:Convex0} (convergence for convex problems):} \label{sec:proof_thm4}
\begin{proof}
First, by the fundamental Theorem of calculus, we have the following identity
\begin{equation*}
    \nabla \mathcal L(\theta) = \nabla \mathcal L(\theta) - \nabla \mathcal L(\theta^*) = \mathcal H_{\theta} (\theta - \theta^*),
\end{equation*}
where $\mathcal H_\theta \in \mathbb R^{d\times d}$ is obtained by integrating $\nabla^2 \mathcal L$ along the path from $\theta^*$ to $\theta$. More precisely, we have
\begin{equation*}
    \mathcal H_\theta = \int_{t=0}^1 \nabla^2 \mathcal L(\theta_t) dt,
\end{equation*}
where $\theta_t = t\theta + (1-t)\theta^*$. Given Assumption~\ref{asm:ms-spectrum} for $\nabla^2 \mathcal L(\theta)$, using the triangle inequality for integral, we can deduce that $\mathcal H_\theta$ also satisfies Assumption~\ref{asm:ms-spectrum} for any $\theta$. 
As a result,
\begin{equation}\label{eq:viteration}
    \theta^k_{j,s+1} - \theta^* = \theta^k_{j,s} - \theta^* - \eta_j \nabla \mathcal L(\theta^k_{j,s}) = (\vec I_d - \eta_j \mathcal H^k_{j,s})( \theta^k_{j,s} - \theta^*).
\end{equation}
The idea of the remaining proof is to decompose the error term into different orthogonal spaces, and show the error will be bounded on each of the space. To this end, let us denote the component of the error in the $i$-th space by 
$$
v_i^{j,s} := \Pi_i \left(\theta^k_{j,s} - \theta^*\right),
$$
where $\Pi_i$ is the row-orthogonal matrix defined in~\Cref{eqn:Pi}. Then, we apply $\Pi_i$ to \Cref{eq:viteration} to get the error in the $i$-th space for the next step, we obtain
\begin{equation}
    \begin{aligned}
        \left\|v_i^{j,s+1} \right\| = &\left\| \Pi_i (\mathbf I_d - \eta_j \mathcal H^k_{j,s}) \left(\sum_{\ell=1}^m \Pi^T_\ell v_\ell^{j,s} \right)  \right\| \\
        \le &\left\|\left(\mathbf I_{d_i} - \eta_j \Pi_i \mathcal H^k_{j,s}\Pi_i^T \right)v_i^{j,s}\right\| + \sum_{\ell \neq i} \eta_j \left\|\Pi_i \mathcal H^k_{j,s} \Pi_\ell^T v_\ell^{j,s}\right\| \\
        \le &\left\|\left(\mathbf I_{d_i} - \eta_j \Pi_i \mathcal H^k_{j,s}\Pi_i^T \right)\right\|\left\|v_i^{j,s}\right\| +\delta \eta_j 
         \sum_{\ell \neq i} \left\| \Pi_\ell (\theta_{j,s}^k - \theta^*) \right\|\\
         \le & \left\|\left(\mathbf I_{d_i} - \eta_j \Pi_i \mathcal H^k_{j,s}\Pi_i^T \right)\right\|\left\|v_i^{j,s}\right\| +\delta \eta_j 
         \left\| \theta_{j,s}^k - \theta^*\right\| \\
        \le &\begin{cases}
         \left(1-R_i^j \kappa_i^{-1}\right)\left\|v_i^{j,s}\right\| + \delta \eta_j C    \quad &\text{if } j \le i,\\
         \left|1-R_i^j\right|\left\|v_i^{j,s}\right\| + \delta \eta_j C     \quad &\text{if } j > i.
        \end{cases}.
    \end{aligned}
\end{equation}

By repeating the above process from $v_i^{j,n_j}$ to $v_i^{j,0}$, we have
\begin{equation}\label{eq:vijbound}
\begin{aligned}
   \left\|v_i^{j,n_j} \right\|  \le &\begin{cases}
       \left(1-R_i^j \kappa_i^{-1}\right)^{n_j}\left\|v_i^{j,0}\right\| + \delta \eta_j C E_i^j   \quad &\text{if } j \le i,\\
         \left|1-R_i^j\right|^{n_j}\left\|v_i^{j,0}\right\| + \delta \eta_j C E_i^j \quad &\text{if } j > i,
   \end{cases} \\
   = &\begin{cases}
       \left(1-R_i^j \kappa_i^{-1}\right)^{n_j}\left\|v_i^{j+1,n_{j+1}}\right\| + \delta \eta_j C E_i^j   \quad &\text{if } j \le i,\\
         \left|1-R_i^j\right|^{n_j}\left\|v_i^{j+1,n_{j+1}}\right\| + \delta \eta_j C E_i^j \quad &\text{if } j > i.
\end{cases} .
\end{aligned}
\end{equation}
since $\theta^k_{j,0} = \theta^k_{j+1,n_{j+1}}$, where
\begin{equation}
    E_i^j = \begin{cases}
         \sum_{s=0}^{n_j-1} \left(1-R_i^j\kappa_i^{-1}\right)^s \quad &\text{if } j \le i, \\
         \sum_{s=0}^{n_j-1} \left|1-R_i^j\right|^s \quad &\text{if } j > i.
    \end{cases}.
\end{equation}
By re-applying the inequality in \Cref{eq:vijbound} repeatedly on $\left\|v_i^{1,n_{1}}\right\|$ we have
\begin{equation}
    \left\|v_i^{1,n_1}\right\| \le c_i \|v_i^{m,0}\| + \delta C \sum_{j=0}^{m-1}\eta_{j+1}C_i^j E_i^j ,
\end{equation} where
\begin{equation}
    c_i = \prod_{j=1}^{i} \left(1 - R_i^j\kappa_{i}^{-1}\right)^{n_j}   \prod_{j=i+1}^m \left|1- R^j_i \right|^{n_j},
\end{equation}
matched with the definition given in \Cref{eq:defci} for all $i=1:m$ and 
\begin{equation}
    C_i^j = \begin{cases}
        \prod_{s=1}^{j} \left(1 - R_i^s\kappa_{i}^{-1}\right)^{n_s} \quad &\text{if } j \le i,\\
       \prod_{s=1}^{i} \left(1 - R_i^s\kappa_{i}^{-1}\right)^{n_s} \prod_{s=i+1}^j \left|1- R^s_i \right|^{n_s} \quad &\text{if } j > i.
    \end{cases}.
\end{equation}
Adopting the notation $\theta^k = \theta^{k}_{1,n_1}$, for a fixed $k$, we have: 
\begin{equation}
\begin{aligned}
    \left\|\theta^k - \theta^{*}\right\| 
    = &\sum_{i=1}^m \left\|\Pi_i(\theta^k - \theta^{*})\right\| 
    = \sum_{i=1}^m \left\|v_i^{1,n_1}\right\| \\
    \le &\sum_{i=1}^m c_i \|v_i^{m,0}\| + \delta C  \sum_{i=1}^m 1 \times \sum_{j=0}^{m-1}C_i^jE_i^j \\
    \le &\max_{i=1:m} \{c_i\} \sum_{i=1}^m \|v_i^{m,0}\| + \delta m C \max_{i=1:m}\left\{\sum_{j=0}^{m-1}C_i^jE_i^j\right\} \\
    = &\max_{i=1:m} \{c_i\} \left\|\theta^{k-1} - \theta^*\right\| + \delta m C \max_{i=1:m}\left\{\sum_{j=0}^{m-1}C_i^jE_i^j\right\}
\end{aligned}
\end{equation}
since $\theta^k_{m,0} = \theta^{k-1}$. 
Recalling~\Cref{eq:ciorder} in the proof of Theorem~\ref{thm:Sm} in Appendix~\ref{sec:proofSm}, we have
\begin{equation}
    \max_{i=1:m} \{c_i\} = c_m = \prod_{j=1}^{m} \left(1 - R_m^j\kappa_{m}^{-1}\right)^{n_j},
\end{equation}
if $n_{m-1}, \cdots, n_2, n_1$ satisfy the condition in \Cref{eq:nicondition}. This finishes the proof for Theorem~\ref{thm:ConvexConvergence}.
\end{proof}

\section{Numerical example for linear regression with multiscale data}\label{appendix_linearexample}
First, the linear regression problem depicted in Figure~\ref{figs:MrGDlinear} is represented as
$$
 \min_{w} \frac{1}{N}\sum_{i=1}^N (w \cdot \vec{x}_i - g_i)^2 \Leftrightarrow A w = b,
$$
where $A = \frac{1}{N}\sum_{i=1}^N\vec{x}_i \vec{x}_i^T \in \mathbb{R}^{100\times 100}$ and $b =  \frac{2}{N}\sum_{i=1}^N g_i \vec{x}_i \in \mathbb{R}^{100}$. In the numerical results presented in Figure~\ref{figs:MrGDlinear}, we record the residual $\|Aw^k - b\|$ on the vertical axis. For the GD method, $k$ represents the number of GD steps. For MrGD, $k$ denotes the total number of iterations, i.e., including both inner and outer iterations.

\paragraph{Two-Scale Problems.}
We sample the data $\vec{x}_i = (\vec{x}_i^0, \sqrt{\varepsilon} \vec{x}_i^1) \in \mathbb{R}^{d_0+d_1}$, where $d_0= 80$ and $d_1=20$. Specifically, we set $\varepsilon=0.001$, sample $g_i \sim N(0,1)$, and sample $\vec{x}_i^k \sim N(0,I_{d_k})$ for $k=0,1$ and $i=1:N$ with $N=10^{4}$. For this case, we can approximately have that $r \approx \varepsilon = 0.001$ and the global condition number of $A$ is around $r^{-1}$. In the MrGD algorithm, we set $\eta_0 = 0.5$, $\eta_1 = \frac{1}{2r}$, and $n_1 = 15$.

\paragraph{Three-Scale Problems.}
We sample the data $\vec{x}_i = (\vec{x}_i^0, \sqrt{\varepsilon} \vec{x}_i^1, \varepsilon \vec{x}_i^2) \in \mathbb{R}^{d_0+d_1+d_2}$, where $d_0= 60$, $d_1=20$, and $d_2=20$. Specifically, we set $\varepsilon=0.1$, sample $g_i \sim N(0,1)$, and sample $\vec{x}_i^k \sim N(0,I_{d_k})$ for $k=0,1,2$ and $i=1:N$ with $N=10^{4}$. For this case, we can approximately have that $r \approx \varepsilon = 0.1$ and the global condition number of $A$ is around $r^{-2}$. In the MrGD algorithm, we set $\eta_0 = 0.5$, $\eta_1 = \frac{1}{2r}$, and $\eta_2 = \frac{1}{2r^2}$ with $n_1 = 3$ and $n_2 = 15$.

\section{CIFAR MNIST Comparison}
\begin{figure}[ht]
\vskip 0.2in
\begin{center}
\centerline{\includegraphics[width=0.45\columnwidth]{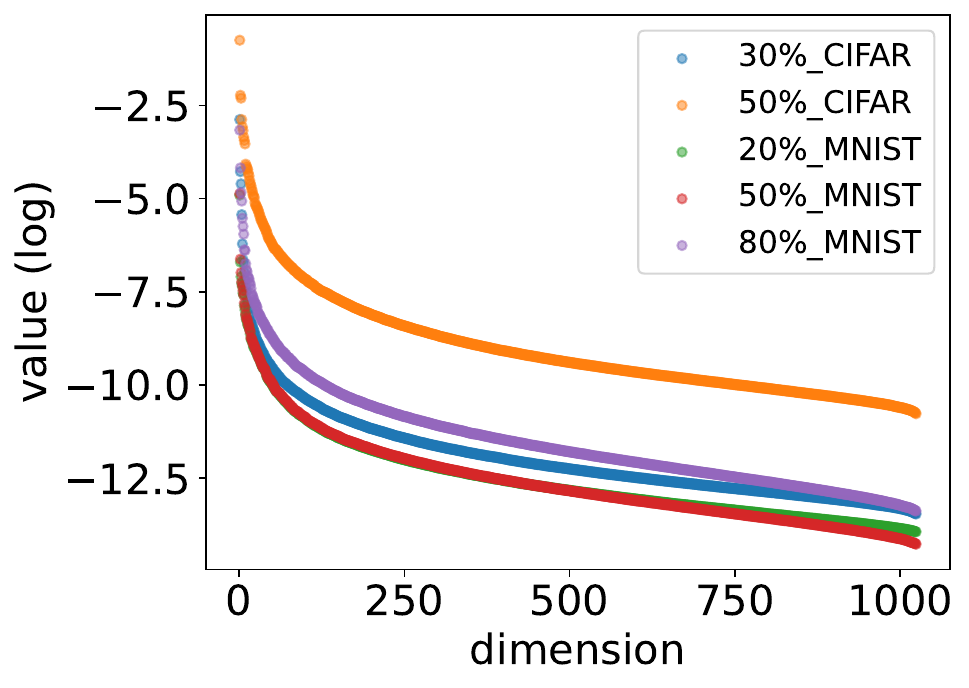}}
\caption{Comparison of eigenvalues (sorted by magnitudes) of the Hessian in the second hidden layer, under the natural log scale. The CIFAR models are those from~\Cref{sec:NN_multiscale}. The MNIST model is also a $3$-layer MLP of sizes $784$-$1024$-$128$-$10$, identical to the CIFAR model except for the input dimension. The MNIST model is also trained under the same setting: full gradient descent, fixed learning rate $0.1$, and cross-entropy loss. Three stages in training of the MNIST model with test accuracy $20\%$, $50\%$, and $80\%$, respectively are presented.  All models in both the CIFAR and MNIST exhibit similar decaying property.}
\label{figs:cifar_mnist_w2}
\end{center}
\vskip -0.2in
\end{figure}



\end{document}